\documentclass[fleqn,10pt]{wlscirep}
\usepackage[utf8]{inputenc}
\usepackage{colortbl} 
\usepackage{algorithm}
\usepackage{algpseudocode}
\usepackage{subcaption}
\usepackage[T1]{fontenc}
\usepackage{tikz}
\usetikzlibrary{shapes, arrows, positioning}
\usepackage{hhline}
\usepackage{url}
\usepackage{lineno,hyperref}
\modulolinenumbers[5]
\usepackage{hhline}
\usepackage{url}
\usepackage{comment}
\usepackage{threeparttable}
\usepackage{textcomp}
\usepackage{parskip}
\usepackage{hhline}
\usepackage{adjustbox}
\usepackage{fixfoot}
\usepackage{lscape} 
\usepackage{booktabs} 
\usepackage{color,soul}
\usepackage{tabularx}
\usepackage{adjustbox}
\usepackage{multirow}
\usepackage{colortbl} 
\definecolor{lightgray}{gray}{0.9}
\definecolor{white}{gray}{1.0}
\definecolor{RED}{RGB}{50 100 200}
\usepackage{afterpage}
\tikzstyle{small} = [draw, rectangle, minimum width=2cm, minimum height=1cm, text centered]
\tikzstyle{big} = [draw, rectangle, minimum width=4cm, minimum height=1cm, text centered]
\title{Infusing clinical knowledge into tokenisers for language models}

\author[1,*]{Abul Hasan}
\author[1]{Jinge Wu}
\author[1]{Quang Ngoc Nguyen}
\author[2]{Salomé Andres}
\author[2]{Imane Guellil}
\author[2]{Huayu Zhang}
\author[2]{Arlene Casey}
\author[2,3,4]{Beatrice Alex}
\author[2]{Bruce Guthrie}
\author[1,2,5,*]{Honghan Wu}
\affil[1]{The UCL Institute of Health Informatics, 222 Euston Road, London, NW1 2DA }
\affil[2]{Advanced Care Research Centre, The Usher Institute, College of Medicine and Veterinary Medicine,
University of Edinburgh, Edinburgh, United Kingdom}
\affil[3]{School of Literatures, Languages and Cultures, University of Edinburgh, Edinburgh,
United Kingdom}
\affil[4]{Edinburgh Futures Institute, University of Edinburgh, Edinburgh, United Kingdom}
\affil[5]{School of Health and Wellbeing, University of Glasgow, Glasgow, United Kingdom}

\affil[*]{corresponding.a.kalam@ucl.ac.uk, honghan.wu@ucl.ac.uk}

\begin{abstract}
This study introduces a novel knowledge enhanced tokenisation mechanism, K-Tokeniser, for clinical text processing. Technically, at initialisation stage, K-Tokeniser populates global representations of tokens based on semantic types of domain concepts (such as drugs or diseases) from either a domain ontology like Unified Medical Language System or the training data of the task related corpus. At training or inference stage, sentence level localised context will be utilised for choosing the optimal global token representation to realise the semantic-based tokenisation. To avoid pretraining using the new tokeniser, an embedding initialisation approach is proposed to generate representations for new tokens. Using three transformer-based language models, a comprehensive set of experiments are conducted on four real-world datasets for evaluating K-Tokeniser in a wide range of clinical text analytics tasks including clinical concept and relation extraction, automated clinical coding, clinical phenotype identification, and clinical research article classification. Overall, our models demonstrate consistent improvements over their counterparts in all tasks. In particular, substantial improvements are observed in the automated clinical coding task with 13\% increase on Micro $F_1$ score. Furthermore, K-Tokeniser also shows significant capacities in facilitating quicker converge of language models. Specifically, using K-Tokeniser, the language models would only require 50\% of the training data to achieve the best performance of the baseline tokeniser using all training data in the concept extraction task and less than 20\% of the data for the automated coding task. It is worth mentioning that all these improvements require no pre-training process, making the approach generalisable.

\end{abstract}
\begin{document}
\flushbottom
\maketitle
\section{Introduction}
\label{sec:01}
All contemporary language models (LMs), such as BERT models \cite{devlin2019bert} or OpenAI's GPT-4 \cite{brown2020language, achiam2023gpt}, are built upon the Transformer \cite{vaswani2017attention} architecture, employing a tokeniser as their initial text-processing block. The tokeniser is a foundational building block of LMs which determines how text is represented when input to a model. While recent advancements in clinical {\em Natural Language Processing (NLP)} have focused on enhancing architectures and gathering more extensive data, little attention has been directed toward this crucial step of tokenisation \cite{yehezkel2023incorporating}. 
For tokenisation, one of the challenges is that a tokeniser's vocabulary inevitably lacks coverage of all words in a clinical dataset. For example, ClinicalBERT \cite{alsentzer2019publicly}, a BERT-based pre-trained model for the clinical domain, has a vocabulary of around 30,000 words. 
When ClinicalBERT encounters a word not in its vocabulary, e.g., \texttt{meropenem} (a drug name), the model's tokeniser segments it into a list of substrings, called {\em subwords}. In this case, the generated subwords from the tokeniser are \texttt{me, \#\#rop, \#\#ene, \#\#m}. However, clearly, there are many other combinations of subwords for this word, like \texttt{mero, \#\#penem} or \texttt{mero, \#\#pen, \#\#em}. Some might seem more natural to human eyes and different people might have different preferences. However, the true question (that this paper focuses on) is which option would best facilitate the capabilities of the language model in doing downstream tasks.   

Efforts have been made to acquire specialised biomedical corpus-specific vocabularies, as seen in PubMedBERT \cite{gu2021domain} (sometimes referred to as BiomedBERT) and the Gatarton \cite{yang2022large} model. However, the PubMedBERT and Gatarton models require the complete retraining of the BERT model from its initial stages, as opposed to ClinicalBERT and BioBERT \cite{lee2020biobert}, which initiate with weights from the original BERT model. Pre-training a model from scratch, as required by PubMedBERT and Gatarton, is expensive due to the significant computational resources and time needed, accompanied with the potential loss of general knowledge learned in the pre-trained weights of the original BERT model. 
Most of all, the main limitation of the pretraining approach resides in the fact that it treats all words equally when generating subwords. It overlooks the fact that different clinical concepts tend to have distinct subword patterns, for example \emph{-itis, -pathy, -penia} are common in diseases and \emph{anti-, beta-, -ol, -tine} are frequent in drugs. Such difference has yet been leveraged in improving LM's capabilities in learning and/or inference. 
Discovering subwords as part of tokenisation that emulate clinically significant patterns (akin to linguistically meaningful units in natural language), is both a technical requirements and an essential component in constructing clinical LMs. 
However, expanding a tokeniser's vocabulary to include subwords from a clinical ontology poses several noteworthy challenges. Firstly, the expansion procedure requires a formal approach to derive subwords from an ontology or training data. Secondly, the random initialisation of embeddings for the expanded vocabulary within a pre-trained LM  does not produce improved performance \cite{tai2020exbert}. Thirdly, the vocabulary expansion has the  potential to alter the count of generated subwords per word, known as {\em fertility \cite{rust2021good}}, which affects the length of sequences/sentences, consequently influencing the overall performance of a pre-trained model.  In this study, we propose a novel framework called K-Tokeniser designed specifically for clinical text processing, which addresses the aforementioned challenges.
Our contributions are as follows:
\begin{enumerate}
    \item We introduce a novel tokenisation mechanism, K-Tokenisation, that expands the vocabulary of a baseline tokeniser by deriving global character representations leveraging drug and symptom related concepts extracted from either a domain ontology, such as the Unified Medical Language System (UMLS) \cite{bodenreider2004unified}, or a task-specific corpus, such as MIMIC-III \cite{johnson2016mimic}.
    \item We propose a simple yet effective method for initialising embeddings of new subwords to transfer knowledge from a pre-trained model, utilising its existing vocabulary.
    \item We integrate two optimisation objectives into the K-Tokeniser at the inference stage, aiming to discover the optimal subword representation by considering both global semantic representations of medical concepts and localised context at the sentence level.
\end{enumerate}


\par  
\section{Overview of Tokenisation and K-Tokeniser}
\label{sec:02}
\subsection{Tokenisation Background}
To learn a a set of predefined vocabulary from a corpus, a compression algorithm like Byte Pair Encoding (BPE) \cite{sennrich2016neural} or its variant is used. In recent advancements such as the GPT-3 model \cite{brown2020language}, BPE serves as the main mechanism for vocabulary generation. Conversely, models based on BERT utilise a variant of the WordPieceModel \cite{schuster2012japanese}. In the case of BPE, a vocabulary of predefined size is learned by finding the most frequent characters in pairs and merging them until the size is reached. In contrast, WordPieceModel and UnigramLM \cite{kudo2018sentencepiece} gradually remove subword units from a starting vocabulary that contains far more subword units than are desired \cite{mielke2021between}.  Based on information theory, these frequency-driven methods represent data compression strategies aimed at minimising entropy, thereby rendering the resultant corpus more conducive to learning and prediction \cite{xu2021vocabulary}. The vocabulary learned during the pre-training phase remains fixed in a subsequent fine tuning phase. In both pre-training and fine-tuning phase, textual data is typically pre-processed into sentences or sequences of words and subwords. It is possible to process a word into multiple types of subwords by segmenting it into various ways. This methodology is already adopted in neural machine translation systems\cite{kudo2018subword} to enhance the robustness of the NLP model. For example, FLOTA \cite{hofmann2022embarrassingly} algorithm processes \texttt{undesirable} as \texttt{un,\#\#desirable} in contrast to the subwords  \texttt{und, \#\#es, \#\#ira, \#\#ble} produced by the 
 {\em BERT Tokeniser} demonstrating improvements in some downstream tasks. 
 \par There have been efforts to augment this fixed vocabulary during the fine-tuning stage by integrating new categories of lexical items and subword units \cite{tai2020exbert, hong2021avocado, mosin2023fine}, but these methodologies often suffer from over fitting issues and exhibit a lack in generalisation capability, notably within the clinical domain \cite{tai2020exbert}. According to the findings presented by the authors of FLOTA \cite{hofmann2022embarrassingly}, it has been observed that the subwords generated by the BPE algorithm and its variants may not consistently encapsulate meaningful word segmentation and exhibit limited morphological coherence \cite{mielke2021between, bostrom2020byte, xue2022byt5}. Moreover, they may completely fail to capture some salient character patterns found in a certain type of semantic category in an ontology such as UMLS \cite{bodenreider2004unified}. For example, within the Clinical Drug semantic categories of UMLS, noteworthy examples like \texttt{atenolol} and \texttt{timolol} share the suffix \texttt{olol}. The BERT Tokeniser fails to capture this similarity, segmenting these terms into subwords such as \texttt{ate, \#\#no, \#\#lo, \#\#l} and \texttt{t, \#\#imo, \#\#lo, \#\#l} respectively.
 \afterpage{\clearpage
 \begin{figure}
    \centering
    \caption{\textbf{K-Tokenisation framework and study design for its evaluation}}\label{fig:01}
    \begin{subfigure}[b]{\textwidth}
        \centering
        \captionsetup{labelformat=empty}
        \caption{a) A schematic of the K-Tokenisation framework}
        \label{fig:01_a}
        \includegraphics[width=\textwidth]{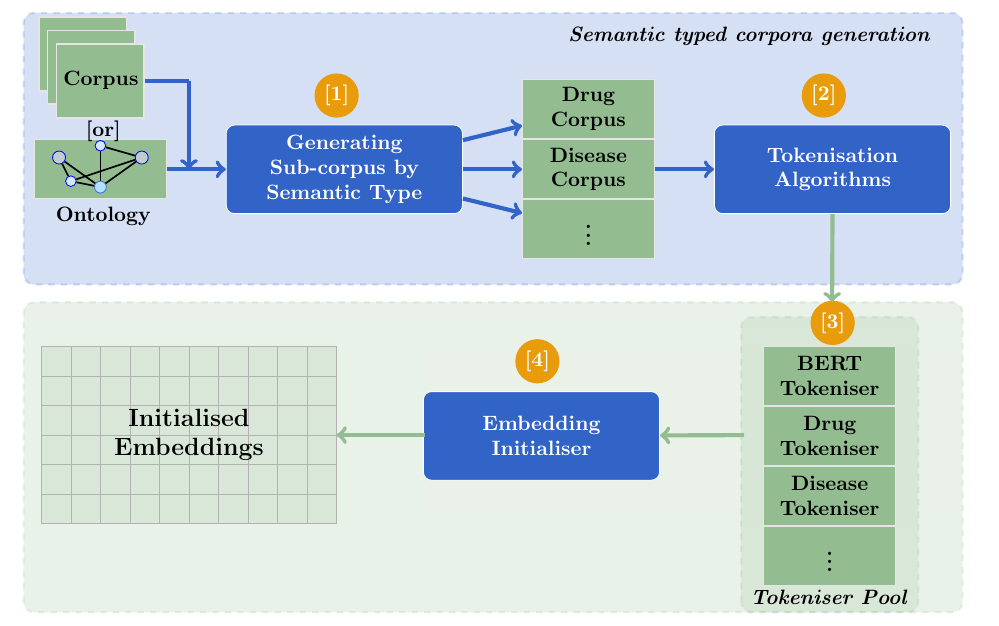}
    \end{subfigure}
    \vspace{1cm}
    \begin{subfigure}[b]{\textwidth}
        \centering
         \captionsetup{labelformat=empty}
        \caption{b) A schematic of the study design for evaluating K-Tokenisation}
        \label{fig:01_b}
        \includegraphics[width=0.98\textwidth]{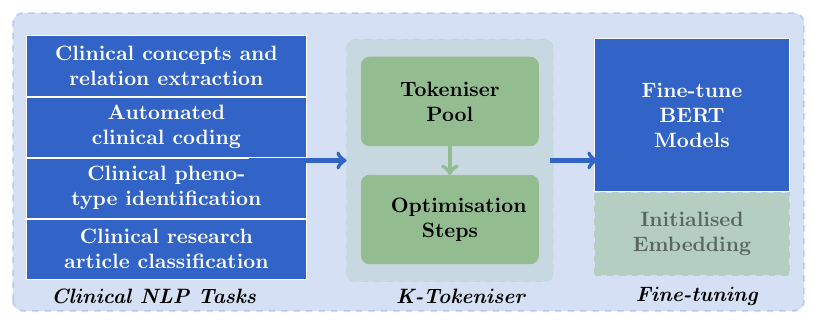}
    \end{subfigure}
    \caption*{\textbf{a)} Four steps of K-Tokeniser framework: [1] Generating Sub-corpus by Semantic Type- identifying and quantifying all sub-strings associated with a specific semantic category (e.g., Clinical Drug, Sign, or Symptom) within a concept vocabulary derived from a domain ontology or corpus. [2] Tokenisation Algorithms- constructing subwords through the utilisation of Sub-corpus, from [1] and by employing two algorithms. [3] Tokeniser Pool-  a pool of semantic type tokenisers which consist of subwords generated from [2]. [4] Embedding Initialiser- initialising transformer embedding layer for new subwords. \textbf{b)} The diagram outlines our study design. Text from four clinical NLP tasks undergoes processing using K-Tokeniser, followed by fine-tuning using an augmented BERT-based pre-trained model, such as ClinicalBERT. }

\end{figure}
\clearpage}
\subsection{A Novel Tokenisation Framework - K-Tokeniser}
In Figure ~\ref{fig:01_a}, we illustrate a schematic of the K-Tokenisation framework. This framework includes the following steps: {\em Generating Sub-corpus by Semantic Type}, {\em Tokenisation Algorithms} for deriving new semantic-based subwords, a {\em Tokeniser Pool} comprising multiple semantic-type tokenisers, and an {\em Embedding Initialiser} for computing representations of new subwords from existing embeddings. For detailed description of these steps, refer to the Methods Section.
\par When the K-Tokeniser encounters a new word at training or inference stage, it generates multiple types of subwords from its tokeniser pool. Using a word by word entropy minimisation objective termed as {\em Word Optimisation}, the K-Tokeniser selects the most appropriate subwords according to its semantic type. As a result of this global optimisation objective, sequences may possess shorter lengths that are not encountered during the pre-training phase. To address this, a localised context based {\em Sequence Optimisation} is established by comparing the fertility of the BERT Tokeniser and K-Tokeniser. A fertility threshold is set below which the K-Tokeniser consistently prioritizes subwords from the BERT Tokeniser. This objective, termed {\em Sequence Optimisation}, ensures consistency with the pre-trained phase. These two steps, collectively called as {\em Optimisation Steps}, are described in detail in the Methods Section. 
\subsection*{Study design}
We aim to enhance clinical and biomedical language models through the integration of K-Tokeniser and investigate whether this new tokenisation approach improves performance across four clinical NLP tasks. In Figure ~\ref{fig:01_b}, we present a schematic of our study design. Text from various datasets undergo preprocessing with K-Tokeniser and subsequent fine-tuning using pre-trained BERT-based models initialised with embeddings from existing ones. The K-Tokeniser incorporates new subwords from three different sources: (i) UMLS ontology, (ii) MIMIC-III hospital discharge summaries, and (iii) PubMed corpus. Upon generating new subwords, we map them to the vocabulary of the BERT Tokeniser to augment a pre-trained model's embedding layer. Thus, this construction allows us to derive augmented pre-trained models from existing models. Models integrated with the K-Tokeniser originating from three sources are identified by prefixing the letter ``K'' at the beginning of their names, followed by the source as a suffix. More specifically, the evaluation involves ClinicalBERT paired with K-ClinicalBERT$_{UMLS}$ and K-ClinicalBERT$_{MIMIC}$, PubMedBERT with K-PubMedBERT$_{UMLS}$ and K-ClinicalBERT$_{PubMed}$, Bioformer with K-Bioformer$_{UMLS}$ and K-Bioformer$_{PubMed}$. We then assess performance and compare it with baseline BERT models. 
\section{Results}
\label{sec:03}
\subsection{Tokeniser Evaluation Design and Datasets}
\par 
Our evaluation includes 4 clinical NLP tasks on the following benchmark datasets: (i) 2018 n2c2 dataset \cite{henry20202018, yang2020identifying, huang2022plm} for clinical concept and relation extraction, (ii) MIMIC-III \cite{johnson2016mimic} (Medical Information Mart for Intensive Care) dataset for automatic coding \cite{mullenbach2018explainable, dong2021explainable},  (iii) RadGraph \cite{jainradgraph} dataset for clinical phenotype identification, and (iii) LitCovid \cite{chen2022multi} corpus for clinical research article classification. The sample sizes of these dataset are shown in Table ~\ref{tab:data}. We utilise publicly available ClinicalBERT \cite{alsentzer2019publicly} model for tasks related to datasets (i), (ii), and (iii), and PubMedBERT \cite{gu2021domain} and Bioformer \cite{chen2022multi, fang2023bioformer} models for the task in (iv). In the following subsections we describe these datasets and their related tasks in detail.
\begin{table}[!h]
     \centering
      \caption{Clinical NLP task and dataset descriptions} \label{tab:data}
    \begin{tabular}{|l|l|l|l|l|l|}
     \hline
     \multirow{2}{*}{NLP task}&\multirow{2}{*}{Dataset name}&\multirow{2}{*}{Text type}& \multicolumn{2}{c|}{No. of samples}&{No. of labels} \\ \cline{4-5} 
     &&&Train&Test& \\ \hline 
     Clinical concepts and relation extraction &n2c2&Discharge summary&303&203&9 \\ \hline
     Automated clinical coding &MIMIC-III&Discharge summary& 47,724&3372&8922 \\ \hline
     Clinical phenotype identification &RadGraph&Radiology report&500&100&4\\ \hline
     Clinical research article classification&LitCovid&PubMed abstracts&24,960&2,500&7 \\ \hline 
     \end{tabular}
     \vspace{0.2cm}
 \caption*{The table includes type of text, number of train and test samples, and labels of each dataset.}
\end{table}
\subsubsection*{Clinical concepts and relation extraction}
Clinical concept extraction is the task of classifying a word or sequence of words (phrases) into semantic categories (e.g., Drug, Strength, Reason), while relation extraction identifies relationships between two concepts. The n2c2 \cite{henry20202018} dataset contains 505 discharge summaries from the MIMIC-III annotated with 83,869 concepts of 9 class labels of which 50,951 and 32,918 concepts are in training and test sets, respectively. Here, the Drug is the dominant concept comprising 32\% of all categories. The Duration, ADE (Adverse Drug Event), and Reason concepts collectively constitute only 11\% of all concepts, with individual proportions of 1\%, 2\%, and 8\%, respectively. This dataset is also annotated with 8 relation types between Drugs and its related attributes (e;g. Drug-ADE denotes a relation between Drug and Adverse Drug Event). In total, there are 59,810 relational pairs of which 36,384 and 23,462 relations are in the training and test set, respectively. Most relation pairs fall in the range of 11-18\%, while ADE-Drug and Duration-Drug pairs comprise only 3\% and 2\%, respectively, of all relation pairs.
\subsubsection*{Automated clinical coding}
Clinical notes such as discharge summaries are accompanied by metadata codes from the {\em International Classification of Diseases (ICD)}. These codes standardize the indication of diagnoses and procedures performed during the encounter, serving various purposes from billing to predictive modelling of the patient state \cite{mullenbach2018explainable}. Manual coding of these notes is labour-intensive and prone to errors, leading to the exploration of automatic coding \cite{de1998hierarchical, mullenbach2018explainable, dong2021explainable}. However, the automation of the coding task is challenging as illustrated in previous studies \cite{dong2022automated} due to a large number of labels (with over 15,000 codes in ICD-9) and long input segments. Our study utilises the ICD-9 benchmark dataset obtained from MIMIC-III and known as CAML\cite{mullenbach2018explainable} which comprises discharge summaries from an intensive care unit. The dataset contains in total 8,922 unique codes as labels assigned to 52,724 discharge summaries; see Table ~\ref{tab:data} for its train and test sizes. We also investigate narrowing down the number of labels to the 50 most frequent classes \cite{mullenbach2018explainable}. In this setting, there are 8,067 documents for training, 1,574 for validation, and 1,730 for testing.
\subsubsection*{Clinical phenotype identification}
Radiography reports consist of unstructured text written by radiologists detailing their findings derived from interpretations of X-ray images. Information extraction from such reports has the potential to improve downstream tasks involving multi-modal models and disease surveillance \cite{jainradgraph}. We compare ClinicalBERT models in extracting phenotype from the Radgraph \cite{jainradgraph} dataset which contains 12,388 entities from 500 radiology reports for training. The test set contains 1,293 and 1,473 entities from 50 reports of MIMIC-CXR \cite{johnson2019mimic} and CheXpert \cite{irvin2019chexpert} datasets, respectively. In this dataset, the Anatomy: Definitely Present (ANAT-DP) label represents 43.3\% while the Observation labels contain the rest. Specifically, Observation Definitely Present (OBS-DP) dominates the other two classes of labels at 40\% of all entities. Observation: Uncertain (OBS-U) and Observation: Definitely Absent (OBS-DA) labels contain 4.7\% and 11.2\% of all entities. Similar distribution patterns are observed in the test datasets, where the ``Anatomy'' label remains prominent, comprising approximately 40\% of all entities.
\subsubsection*{Clinical research article classification}
LitCovid is a curated literature hub that continuously adds new publications related to COVID-19 and SARS-CoV-2 while categorizing them into research topics \cite{chen2021litcovid}. To automate the categorisation of the database into topics, the BioCreative VII LitCovid Track for COVID-19 literature topic annotation task has released a dataset containing seven types of topic annotations for approximately 30,000 articles \cite{chen2022multi}. Among the topics, Prevention, Diagnosis, and Mechanism are the most prevalent consisting of 44.48\%, 24.81\%, and 17.78\%, of all documents, respectively. 

\begin{table}[!h]
    \centering
    \caption{Results of ClinicalBERT models employing BERT Tokeniser and K-Tokeniser on clinical concepts and relation extraction tasks} \label{tab:01}
    \begin{tabular}{|l|l|l|l||l|l|l|l|}
    \hline 
    \multicolumn{4}{|c|}{Clinical concepts extraction}&\multicolumn{4}{c|}{Relation extraction} \\  \hline
    \multirow{2}{*}{Class}&\multirow{2}{*}{ClinicalBERT}&\multicolumn{2}{c||}{K-ClinicalBERT}&\multirow{2}{*}{Relation}&\multirow{2}{*}{ClinicalBERT}&\multicolumn{2}{c|}{K-ClinicalBERT} \\ \cline{3-4}  \cline{7-8}
    &&$_{UMLS}$&$_{MIMIC}$&&&$_{UMLS}$&$_{MIMIC}$ \\ \hline  
    Drug& 95.49$\pm 0.16$ & 95.35$\pm 0.31$ & \textbf{95.67$\pm 0.10$ }&N/A&N/A&N/A&N/A \\ \hline 
    ADE& 54.77$\pm 1.15$ & 55.40$\pm 1.21$ & \textbf{55.42}$\pm 0.98$&ADE-Drug&78.59$\pm 0.21$&\textbf{79.34}$\pm 0.78$&77.78$\pm 0.47$ \\ \hline 
   Dosage& 93.33$\pm 0.16$ & \textbf{93.81}$\pm 0.36$ & 93.70$\pm 0.12$ &Dosage-Drug& 95.95$\pm 0.35$&\textbf{96.22}$\pm 0.14$&96.07$\pm 0.32$ \\  \hline 
   Duration& 70.72$\pm 0.97$ & \textbf{73.06}$\pm 0.70$ & 71.64$\pm 1.30$ &Duration-Drug&91.25$\pm 0.28$&\textbf{91.87}$\pm 0.21$&91.49$\pm 0.48$ \\ \hline 
   Form& 90.78$\pm 0.17$ & \textbf{91.98}$\pm 1.39$ & 91.21$\pm 0.20$ &Form-Drug&\textbf{97.26}$\pm 0.15$&97.24$\pm 0.12$&97.19$\pm 0.12$\\ \hline 
   Frequency& 88.36$\pm 0.39$ & \textbf{89.22}$\pm 1.21$ & 88.64$\pm 0.21$ &Frequency-Drug&97.91$\pm 0.01$&\textbf{98.05}$\pm 0.09$&97.82$\pm 0.09$ \\ \hline 
   Reason& 61.75$\pm 1.25$ & \textbf{63.49}$\pm 0.59$ & 63.19$\pm 0.50$ &Reason-Drug&\textbf{85.23}$\pm 0.38$&85.07$\pm 0.25$&84.76$\pm 0.30$\\ \hline  
   Route& 94.02$\pm 0.79$ & 94.02$\pm 0.69$ & \textbf{94.52}$\pm 0.45$ &Route-Drug&\textbf{96.48}$\pm 0.37$&96.17$\pm 0.20$&96.01$\pm 0.13$ \\ \hline 
  Strength & 96.70$\pm 0.18$ & 96.71$\pm 0.11$ & \textbf{96.76}$\pm 0.02$ &Strength-Drug&97.57$\pm 0.16$&\textbf{97.92}$\pm 0.10$&97.72$\pm 0.17$\\ \hline \hline
 Micro $F_1$& 91.39$\pm 0.05$ & \underline{\textbf{91.73$\pm 0.16$}} & \textbf{91.71$\pm 0.02$} &Micro $F_1$&\textbf{94.62$\pm 0.22$}&\underline{\textbf{94.81$\pm 0.08$}}&\textbf{94.62$\pm 0.15$}\\ \hline 
   \end{tabular}
    \vspace{0.2cm}
    \caption*{The models utilising K-Tokeniser, constructed from UMLS ontology and MIMIC-III corpus, are denoted as K-ClinicalBERT$_{UMLS}$ and K-ClinicalBERT$_{MIMIC}$, respectively. The upper 9 rows of the table are $F_1$ scores for individual concept and relation classes and the bottom row is for average Micro $F_1$ scores. The scores presented here are obtained by using K-Tokeniser fertility threshold of 0. Standard deviations are computed from the average performance of the top models across three independent runs. Boldface in the upper nine rows indicates the best performance achieved by one of the three models. For the final row,  bold and underline indicate the best performance, while bold alone signifies the next best Micro $F_1$ score. Overall precision and recall results are provided in Supplementary Table 1 of \href{./supplementary_files.pdf}{Supplementary Files}. }
\end{table}
\begin{table}[!h]
    \centering
    \caption{Results of ClinicalBERT models employing BERT Tokeniser and K-Tokeniser on the automated clinical coding classification task}\label{tab:02}
    \begin{tabular}{|l||l|l|l||l|l|l|}
    \hline 
    \multicolumn{7}{|c|}{Top 50 labels} \\ \hline
    &\multicolumn{3}{c||}{Macro scores}&\multicolumn{3}{c|}{Micro scores} \\ \cline{2-4} \cline{5-7}
    \multirow{2}{*}{Score}&\multirow{2}{*}{ClinicalBERT}&\multicolumn{2}{c||}{K-ClinicalBERT}&\multirow{2}{*}{ClinicalBERT}&\multicolumn{2}{c|}{K-ClinicalBERT} \\ \cline{3-4} \cline{6-7}
    &&$_{UMLS}$&$_{MIMIC}$&&$_{UMLS}$&$_{MIMIC}$ \\ \hline 
    Precision &52.49$\pm 0.90$&\textbf{61.92$\pm 0.45$}& \underline{\textbf{62.47$\pm 0.31$}}& 61.15$\pm 1.46$ &\textbf{67.82$\pm 0.22$}&\underline{\textbf{68.45$\pm 0.53$}} \\ \hline 
    Recall& 40.43$\pm 1.54$ &\textbf{59.86$\pm 0.22$}&\underline{\textbf{59.63$\pm 0.99$}}&46.86$\pm 1.20$ &\textbf{65.04$\pm 0.17$}&\underline{\textbf{64.57$\pm 0.97$}}\\ \hline
    $F_1$&45.64$\pm 0.70$&\textbf{60.87$\pm 0.16$}&\underline{\textbf{61.01$\pm 0.39$}}&53.03$\pm 0.24$&\textbf{66.40$\pm 0.09$}&\underline{\textbf{66.44$\pm 0.38$}}\\  \hline
    \multicolumn{7}{|c|}{Full label set} \\ \hline
    Precision&4.81$\pm 0.19$&\underline{\textbf{7.50$\pm 0.12$}}&\textbf{7.37$\pm 0.26$}&65.42$\pm0.99$& \textbf{70.27$\pm 0.54$}&\underline{\textbf{70.55$\pm 0.66$}} \\ \hline
    Recall&3.27$\pm 0.16$&\underline{\textbf{6.01$\pm 0.17$}}&\textbf{5.75$\pm0.26$}&30.68$\pm 1.49$&\underline{\textbf{48.74$\pm 0.56$}}& \textbf{48.17$\pm 0.61$}\\\hline 
    $F_1$&3.89$\pm 0.17$&\underline{\textbf{6.67$\pm 0.15$}}&\textbf{6.46$\pm 0.27$}&41.75$\pm 1.59$ &\underline{\textbf{57.55$\pm 0.21$}}&\textbf{57.24$\pm 0.25$} \\ \hline 
    \end{tabular}
    \caption*{The models utilising K-Tokeniser, constructed from UMLS ontology and MIMIC-III corpus, are denoted as K-ClinicalBERT$_{UMLS}$ and K-ClinicalBERT$_{MIMIC}$, respectively. The scores presented here are obtained by using K-Tokeniser fertility threshold of 0. Standard deviations are computed from the average performance of the top models across three independent runs. Bold and underline indicate the best performance, while bold alone signifies the next best. Supplementary Table 4 in \href{./supplementary_files.pdf}{Supplementary Files}  presents $F_1$ scores for individual labels (i.e., ICD-9 codes) for the top 50 labels.} 
\end{table}
\begin{table}[!h]
    \centering
    \caption{Results of Clinical BERT models employing BERT Tokeniser and K-Tokeniser on the clinical phenotype identification task} \label{tab:03}
    \begin{tabular}{|l||l|l|l||l|l|l|}
    \hline 
    \multicolumn{4}{|c||}{MIMIC-CXR}&\multicolumn{3}{c|}{CheXpert} \\ \hline 
    Class & ClinicalBERT & \multicolumn{2}{c||}{K-ClinicalBERT} & ClinicalBERT & \multicolumn{2}{c|}{K-ClinicalBERT} \\  \cline{3-4} \cline{6-7} 
    & & $_{UMLS}$ & $_{MIMIC}$ & & $_{UMLS}$ & $_{MIMIC}$ \\ \hline 
    ANAT-DP & 96.15 $\pm 0.27$ & 96.49 $\pm 0.19$ & \textbf{96.64} $\pm 0.20$ &89.04 $\pm  0.13 $ & \textbf{89.40} $\pm 0.54 $ & 89.17 $\pm 0.69$  \\ \hline
    OBS-DA & 97.07 $\pm 0.11$ & \textbf{97.11} $\pm 0.10$ & 97.02 $\pm 0.05$  & 96.71 $\pm  0.84 $ & 96.50 $\pm 0.20 $ & \textbf{96.90} $\pm 0.70 $ \\ \hline
    OBS-DP & 86.66 $\pm 0.63$ & \textbf{86.80} $\pm 0.49$ & 86.18 $\pm 0.42$ & 82.12 $\pm  0.85 $ & \textbf{82.99} $\pm 1.09 $ & 82.12 $\pm 0.93 $ \\ \hline
    OBS-U & 73.61 $\pm 1.80$ & \textbf{80.85} $\pm 0.34$ & 73.84 $\pm 2.51$ & 74.89 $\pm  1.09 $ & \textbf{75.26} $\pm 1.2 $ & 71.90 $\pm 1.1 $ \\ \hline
    Micro $F_1$ & 92.74 $\pm 0.29$ & \underline{\textbf{93.18 $\pm 0.22$}} & \textbf{92.78 $\pm 0.09$}& \textbf{87.18 $\pm  0.47 $} & \underline{\textbf{87.62$\pm 0.23 $}} & 87.07 $\pm 0.38 $ \\ \hline
    
    \end{tabular}
    \vspace{0.2cm}
    \caption*{The models utilising K-Tokeniser, constructed from UMLS ontology and MIMIC-III corpus, are denoted as K-ClinicalBERT$_{UMLS}$ and K-ClinicalBERT$_{MIMIC}$, respectively. The $F_1$ scores presented here are obtained by using K-Tokeniser fertility threshold of 0. Standard deviations are computed from the average performance of the top models across three independent runs. Boldface in the upper 4 rows indicates the best performance achieved by one of the three models. For the final row, bold and underline indicate the best performance, while bold alone signifies the next best. Overall precision and recall results are provided in  Supplementary Table 2 of \href{./supplementary_files.pdf}{Supplementary Files}.} 
\end{table}
 \begin{table}[!h]
 \centering
 \caption{Results of PubMedBERT and Bioformer models employing BERT Tokeniser and K-Tokeniser on the clinical research article classification task } \label{tab:04}
 \begin{tabular}{|l||l|l|l||l|l|l|} \hline
 \multirow{2}{*}{Label} & \multirow{2}{*}{PubMedBERT} & \multicolumn{2}{c||}{K-PubMedBERT} & \multirow{2}{*}{Bioformer} & \multicolumn{2}{c|}{K-Bioformer} \\ \cline{3-4} \cline{6-7}
 &&$_{UMLS}$&$_{PubMed}$&&$_{UMLS}$&$_{PubMed}$ \\ \hline
 Epidemic Forecasting & 76.78 $\pm 0.25 $ & 77.11 $\pm 0.17 $ & \textbf{77.96} $\pm 0.85 $ & 76.05 $\pm 0.84 $ & \textbf{76.19} $\pm 0.98 $ & 74.56 $\pm 0.58 $ \\ \hline
Treatment & 90.79 $\pm 0.08 $ & \textbf{90.98} $\pm 0.18 $ & 90.83 $\pm 0.13 $ & 90.58 $\pm 0.17 $ & 90.74 $\pm 0.18 $ & \textbf{90.75} $\pm 0.21 $ \\ \hline
Prevention & 94.29 $\pm 0.07 $ & \textbf{94.60} $\pm 0.19 $ & 94.39 $\pm 0.05 $ & 94.43 $\pm 0.15 $ & 94.39 $\pm 0.05 $ & \textbf{94.48} $\pm 0.15 $ \\ \hline
Mechanism & \textbf{87.98} $\pm 0.32 $ & 87.74 $\pm 0.31 $ & 87.69 $\pm 0.35 $ & 88.05 $\pm 0.19 $ & 87.60 $\pm 0.18 $ & \textbf{87.76} $\pm 0.24 $\\ \hline
Case Report & \textbf{90.66} $\pm 0.29 $ & 90.32 $\pm 0.09 $ & 90.39 $\pm 0.40 $ & 89.86 $\pm 0.20 $ & 89.73 $\pm 0.13 $ & \textbf{90.09} $\pm 0.09 $\\ \hline
Transmission & 69.10 $\pm 0.90 $ & \textbf{69.88} $\pm 0.96 $ & 69.03 $\pm 0.30 $ & 68.35 $\pm 1.74 $ & \textbf{68.73} $\pm 0.36 $ & 67.84 $\pm 1.75 $\\ \hline
Diagnosis & \textbf{88.65} $\pm 0.14 $ & 88.29 $\pm 0.30 $ & 88.64 $\pm 0.16 $ & 88.30 $\pm 0.12 $ & 88.23 $\pm 0.18 $ & \textbf{88.52} $\pm 0.25 $\\ \hline
 Micro $F_1$&89.77 $\pm 0.07 $ &\underline{\textbf{89.91 $\pm 0.11$}}  & \textbf{89.85 $\pm 0.03 $}& \textbf{89.65} $\pm 0.10 $ &89.54 $\pm 0.09 $ & \underline{\textbf{89.71 $\pm 0.09 $}} \\ \hline 
\end{tabular}
\vspace{0.2cm}
 \caption*{The models utilising K-Tokeniser, constructed from UMLS ontology and PubMed corpus, are denoted as K-PubMedBERT$_{UMLS}$ and K-Biofromer$_{UMLS}$, and  K-PubMedBERT$_{PubMed}$ and K-Biofromer$_{PubMed}$, respectively. The scores presented here are obtained by using K-Tokeniser fertility threshold of 0.065. Standard deviations are computed from the average performance of the top models across three independent runs. Boldface in the upper 7 rows indicates the best performance achieved by one of the three models. In the final row, bold and underline indicate the best performance, while bold alone signifies the next best. Overall precision and recall results are provided
 in Supplementary Table 3 of \href{./supplementary_files.pdf}{Supplementary Files}.} 
 \end{table}
\subsection{Results on Clinical NLP Tasks}
In the following subsections we describe our results on the aforementioned four downstream clinical NLP tasks using the K-Tokeniser and compare it with the baseline BERT Tokeniser. 
\subsubsection*{Clinical concepts and relation extraction - a case on adverse drug reaction}
It can be observed from Table ~\ref{tab:01} that the baseline ClinicalBERT model attains an average $F_1$ score of 91.39\%, while the K-Tokniser models achieve higher performances. Specifically, using the UMLS ontology for vocabulary generation has improved the performances of the K-ClinicalBERT$_{UMLS}$ to $F_1$ score of 91.73\%. At the conceptual level, a marginal reduction in the $F_1$ score is observed specifically in the Drug concept, from 95.49\% to 95.35\%, alongside a minor decline in performance pertaining to the Form concept for the UMLS-based model. The K-ClinicalBERT$_{MIMIC}$ demonstrates improvements across all classes; the model further achieves a slight improvement from 95.49\%  to 95.67\% for recognising Drug concepts. Additionally, noteworthy improvements are observed in ADE detection using the K-Tokenisers. For example, The performance of K-ClinicalBERT$_{MIMIC}$ model reaches 55.40\% compared to 54.77\% achieved by the baseline model; see Table ~\ref{tab:01}. In the case of relation extraction, K-ClinicalBERT$_{UMLS}$ model exhibits minor improvement over the baseline model. Particularly, the ADE-DRUG relation class demonstrates a good improvement, highlighting the K-ClinicalBERT$_{UMLS}$ model's relevance in tasks reliant on knowledge. Despite a slight dip in drug-related performance during concept extraction, this does not adversely impact the proficiency in the relation extraction task. We also observe that the K-Tokeniser models achieve higher performances than the baseline model, particularly in scenarios with lower concept support and a reduced number of unique concepts (e.g., Strength, Duration, Frequency). This phenomenon underscores the increment in the capacity of the models facilitated by the K-Tokeniser. 
\subsubsection*{Automated clinical coding - a case on Intensive Care Unit (using the MIMIC III)}
We investigate the effectiveness of the K-Tokeniser in conjunction with the ClinicalBERT model using a recent state-of-the-art neural architecture named PLM-ICD \cite{huang2022plm}. 
For the top 50 labels, ClinicalBERT achieves a Macro $F_1$ score of 45.64\% (and 3.89\% for the full label set), while both K-ClinicalBERT$_{UMLS}$ and K-ClinicalBERT$_{MIMIC}$ models achieve higher scores for the top 50 labels of 60.87\% (full label set 6.67\%) and 61.01\% (6.46\%), respectively.
Our findings in Table ~\ref{tab:02} reveal that ClinicalBERT attains Micro $F_1$ scores of 53.03\% for the top 50 label set. Conversely, the models built using the K-Tokeniser; i.e. K-ClinicalBERT$_{UMLS}$ and K-ClinicalBERT$_{MIMIC}$ outperform the baseline model by a substantial margin and achieve Micro $F_1$ scores of 66.40\% and 66.44\%, respectively. For the full label set the baseline model reaches 41.75\% whereas K-ClinicalBERT$_{UMLS}$ and K-ClinicalBERT$_{MIMIC}$ achieves 57.55\% and 57.24\%, respectively —close to the state-of-the-art performances that range between 58-60\% according to Dong et al \cite{dong2022automated}. We further discuss the performance improvements of individual labels in the Discussion section of the article.
\subsubsection*{Clinical phenotype identification - a case on radiography (using the RadGraph dataset)}
The results in Table ~\ref{tab:03} show that for the MIMIC-CXR test set, the K-ClinicalBERT$_{MIMIC}$ model outperforms the ClinicalBERT model, achieving a Micro $F_1$ score of 92.78\% instead of 92.74\%. The K-ClinicalBERT$_{UMLS}$ model achieves an even higher score (Micro $F_1$  of 93.18\%) than K-ClinicalBERT$_{MIMIC}$, demonstrating higher performances across most of the phenotype labels on the MIMIC-CXR test set. In the case of the CheXpert test set, K-ClinicalBERT$_{UMLS}$ achieves a higher score than ClinicalBERT, increasing from 87.18\% to 87.62\%, while the K-ClinicalBERT$_{MIMIC}$ model achieves a slightly lower score of 87.07\% compared to the ClinicalBERT model. 
\subsubsection*{Clinical research article classification - a case on COVID-19 (using the LitCovid dataset)}
Following the BioCreative VII LitCovid Track \cite{chen2022multi}, we approach this problem as a multi-class classification task. We compare the performances of the K-Tokeniser employing PubMedBERT and Bioformer models on this dataset. Notably, Bioformer has demonstrated state-of-the-art performance in this track \cite{chen2022multi}. Our evaluation shows that integrating the K-Tokeniser enhances the performance of both PubMedBERT and Bioformer models; see Table ~\ref{tab:04}. The K-PubMedBERT$_{UMLS}$ outperforms both baseline models and achieves a Micro $F_1$ score of 89.91\%. 
Although Bioformer$_{UMLS}$ under performs compared to its baseline counterpart, Bioformer$_{PubMed}$ outperforms the Bioformer, with an increase from 89.65\% to 89.71\% in Micro $F_1$ score. Moreover, the performances of individual labels are enhanced by the K-Tokeniser models; see final two columns of Table ~\ref{tab:04}.
 \afterpage{\clearpage
\begin{figure}[!h]
    \centering
     \caption{\textbf{Effect of variable training size using ontology-based (UMLS) K-Tokeniser models}}\label{fig:02}
    \includegraphics[width=\textwidth]{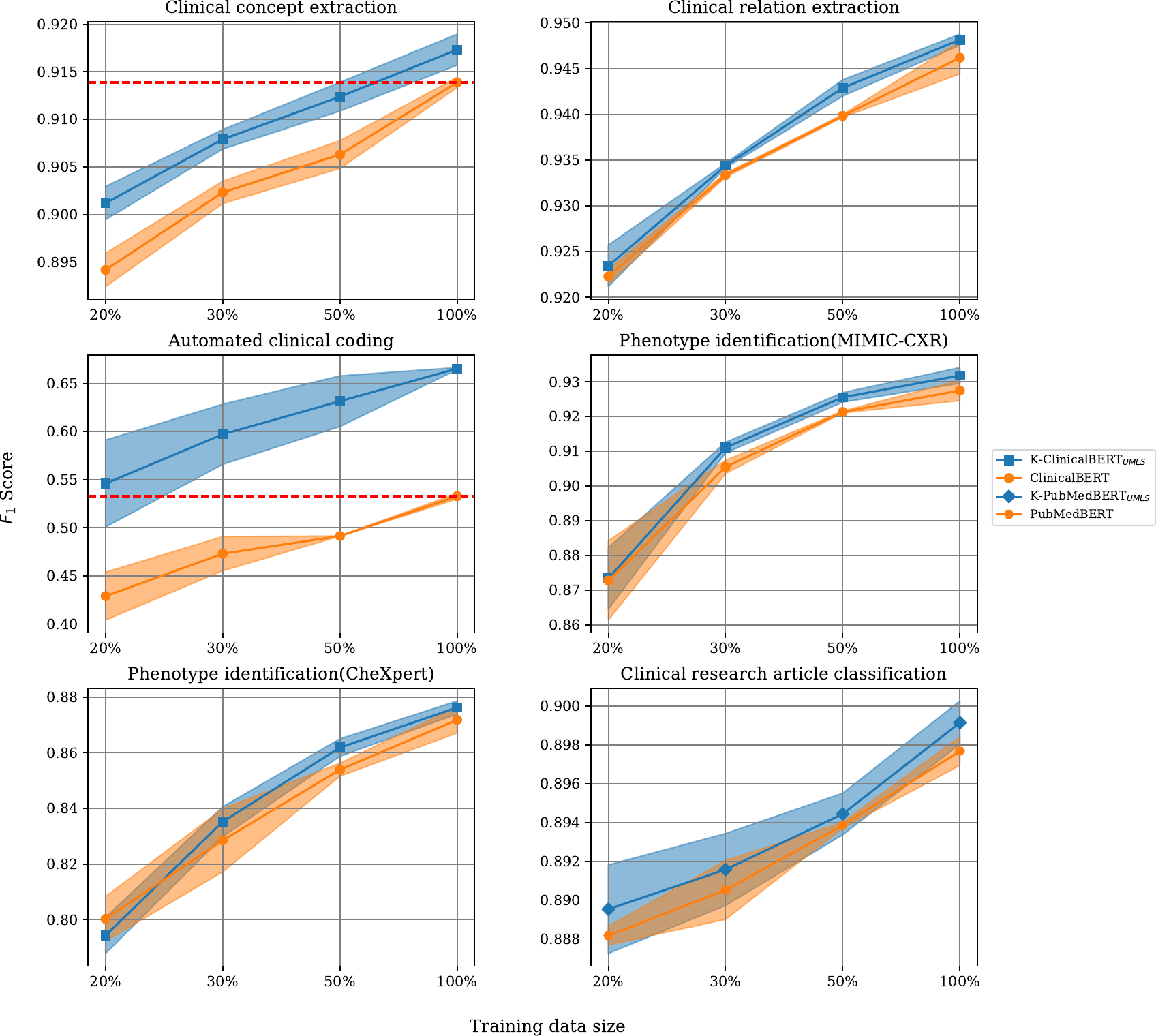}
    \caption*{Comparison of performance between models using the K-Tokeniser from the UMLS ontology and those constructed with corresponding BERT tokenisers across various training sizes. The shaded areas indicate the standard deviation across different partitions and runs. The red dashed lines in the top and middle figures of the left column show the proportion of training data needed for the K-Tokeniser models to match the baseline model's performance. For corpus based tokenisation results see Supplementary Figure 1 of \href{./supplementary_files.pdf}{Supplementary Files} }
  \label{fig:02}
\end{figure}
\clearpage 
}
\subsection{Effect of Variable Training Size}
\par Finally, to compare the performance of K-ClinicalBERT, K-PubMedBERT, and K-Bioformer against ClinicalBERT, PubMedBERT, and Bioformer, respectively, on smaller training set sizes, we conduct an experiment involving the partitioning of the training data of each task into increments of 20\%, 30\%, 50\%, and 100\%. For instance, at 20\% data, we segment the training set into five equal partitions, three partitions for 30\%, and two partitions for 50\%. Furthermore, at 100\% training data, we ran each task 3 times using different seeds. The average $F_1$ score across all partitions of the training set and the corresponding standard deviations are calculated. The results in Figure ~\ref{fig:02} indicate that models constructed using the K-Tokeniser (i.e. K-ClinicalBERT$_{UMLS}$ and K-PubMedBERT$_{UMLS}$) consistently outperform their counterpart baseline models (i.e. ClinicalBERT and PubMedBERT) across all training data sizes across datasets and their corresponding tasks. For the effectiveness of using training data, K-Tokeniser based models demonstrate superior capacity, especially on automated coding tasks (20\% data for K-ClinicalBERT outperforms the baseline using all data) and clinical concept extraction (around 50\% data for K-ClinicalBERT to achieve a performance on par with the baseline using 100\% data). A similar level of improvements are also observed for corpus-based K-ClinicalBERT$_{MIMIC}$; refer to the red dashed lines in the top and middle figures of the left column of Figure ~\ref{fig:02} and Supplementary Figure 1 of \href{./supplementary_files.pdf}{Supplementary Files}.
\section{Discussion}
\label{sec:04}
\par
The K-Tokenisation, when integrated with BERT models, consistently demonstrates performance improvements across all downstream tasks in the clinical domain. Moreover, our analysis reveals that ontology-based tokenisation yields more consistent performances compared to corpus-based tokenisation. This outcome may arise from the presence of spelling mistakes, implicit mentions, and abbreviations within clinical corpora such as MIMIC-III, as identified in a prior research \cite{searle2020experimentals}.  
While the baseline ClinicalBERT models already achieve high performances, and room for improvements are limited, we observe significant enhancements particularly when the baseline performance is lower, as seen in automated clinical coding tasks. Moreover, it requires less training data to achieve the baseline model's accuracy, indicating faster convergence. For instance, in the ICD-9 clinical coding task, models built using the K-Tokeniser achieve the baseline model's accuracy using only 20\% of the training data (see the red dashed line in the middle figures of the left column in Figure ~\ref{fig:02} and Supplementary Figure 1 of the \href{./supplementary_files.pdf}{Supplementary Files}). We posit that these improvements arise from global and local optimisation objectives, in conjunction with the embedding initialiser. Importantly, these steps occur during inference stages, resulting in reduced computing resources without the need for expensive pre-training. Subsequently, we illustrate the progressive improvements achieved by the optimisation steps in conjunction with the embedding initialiser. We then discuss additional experiments aimed at establishing the rationale behind proposing a local context-level optimisation step using the fertility metric. Additionally, we evaluate the efficacy of the optimisation step that involve global concept-level representation. Finally, we provide evidence of the efficacy of semantic type tokenisation by analysing the performance improvement achieved in the clinical coding task, focusing on disease-related ICD-9 coding.
\par Our initial exploration reveals that the expansion of a pre-trained model's vocabulary during the fine-tuning stage does not consistently translate into performance improvements. Consequently, we integrate global(word) and local (sentence) optimisation steps to improve the models. The results presented in Table ~\ref{tab:05} highlight the progressive improvement in model performance across 3 benchmark datasets. Initially, enlarging the vocabulary yields performances marginally lower than those of the baseline BERT models; see first row in Table ~\ref{tab:05}. However, as we update the K-Tokeniser with optimisation objectives, the models' overall $F_1$ score gradually improves. Particularly noteworthy is the substantial performance improvements observed upon initialising transformer embedding layers with average subword embeddings alongside word optimisation procedures; see row 3 of Table ~\ref{tab:05}. We observe performances of all the models constructed using the K-Tokeniser improve across four tasks. Notably, the performance improvement on the ICD-9 coding classification (column (d)) is as much as 13\% over the baseline models. This emphasises that while expanding the vocabulary increases model capacity, it does not necessarily translate to improved performance without embedding layer initialisation of the transformer layer. This, in turn, underscores the importance of ensuring tokenisation procedures generate consistent subwords aligned with the pre-training phase. To address this, we introduce a sentence-level metric termed ``fertility'', measuring the ratio of subwords produced by the K-Tokeniser to those by the BERT Tokeniser. Whenever the fertility exceeds a specified threshold, the K-Tokeniser resorts to subwords proposed by the BERT Tokeniser. Consequently, with a fertility threshold set at 0, the K-Tokeniser consistently defaults to BERT Tokeniser subwords, whereas a threshold of 1 prompts the selection of subwords generated by the optimisation procedure. It can be observed from Table ~\ref{tab:05} that across different fertility thresholds, models' performances vary marginally. However, enhancement of the capacity of the tokeniser with a rule to select a baseline tokeniser always improves the performance, particularly in fine-tuning settings. Next, we discuss how we derive these fertility thresholds using a benchmark dataset.
\begin{table}[!h]
    \centering
    \caption{Ablation study} \label{tab:05}
    \begin{tabular}{|l|l|l||l||l||l|} \hline 
         K-Tokeniser&Fertility&(a)&(b)&(c)&(d) \\ \hline 
         +Random Initialisation&N/A&90.77&94.56& 89.51&53.10 \\
         +Embedding Initialisation&N/A&90.50&94.63&89.54&52.62 \\ \hline 
         \multirow{4}{*}{+Word optimisation}&1.00&91.68&94.88&89.63&66.93\\ 
         &0.06&91.60&94.66&\underline{\textbf{89.91}}& 65.61 \\
         &0.03&91.65&94.78&89.63& 66.57\\
         &0.00&\underline{\textbf{91.73}}&\underline{\textbf{94.82}}&89.40&\underline{\textbf{66.40}} \\ \hline 
    \end{tabular}
    \vspace{0.2cm}
    \caption*{The columns (a), (b), (c), and (d) represent clinical concepts extraction, relation extraction, clinical research article classification, and automated clinical coding tasks. All the experiments are performed using the $UMLS$ based K-Tokeniser models and reported results in Tables ~\ref{tab:01}, ~\ref{tab:02}, ~\ref{tab:04} are bold faced and underlined.} 
\end{table}

\par
\begin{figure}[!htb]
\centering
    \caption{\textbf{Analysis of fertility on the clinical concept extraction tasks using n2c2 dataset}}\label{fig:03}
   \includegraphics[scale=0.58]{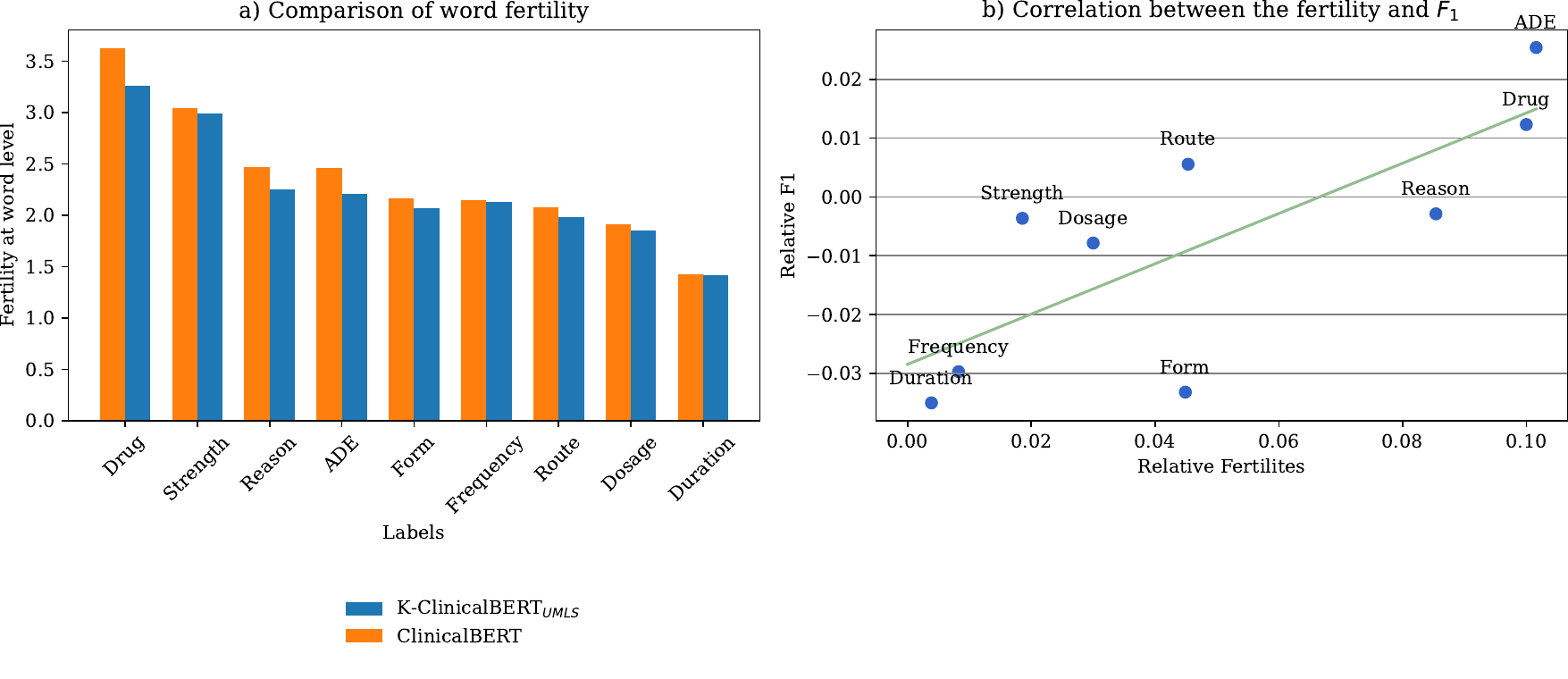}
 \caption*{The results dsiplayed include a) Comparison of word fertility across various classes generated by the K-Tokeniser and the BERT Tokeniser. b) Evaluation of the correlation between the variance in fertility and corresponding $F_1$ scores for each class in models constructed using the two tokenisers.  The K-ClinicalBERT$_{UMLS}$ model is constructed using the K-Tokeniser from the UMLS corpus. }
\end{figure}

 We quantify the number of subwords generated by each category within the n2c2 corpus (i.e. benchmark dataset for clinical concepts and relation extraction) using both the tokenisers. Our analysis reveals consistently lower fertility across all classes when employing the K-Tokeniser; see (a) in Figure ~\ref{fig:03}. This difference is particularly notable in dominant classes like Drug, Reason, and ADE. The K-Tokeniser is designed using semantic categories related to Clinical Drugs and Diseases and Syndrome pertaining on Drug, Reason and ADE class labels, respectively. Building on this observation, we examine the correlations between $F_1$ scores and fertility. Our analysis shows a subtle yet discernible relationship between fertility and $F_1$ scores. Specifically, as the variation in fertility increases, corresponding differences in F1 scores emerge; see (b) in Figure ~\ref{fig:02}. This observation motivates us to set the fertility thresholds at two extreme values 0 and 1, as well as between 0 and 1, specifically at 0.035 and 0.065.
\par We assess the efficacy of the word optimisation step (i.e. selecting subwords based on global character representation of a semantic type) in the context of selecting either the Drug tokeniser or the BERT Tokeniser when it is presented with labelled Drug concepts within the dataset. Our methodology involves extracting all Drug concepts from the n2c2 dataset and analysing the subwords generated by the K-Tokeniser. We calculate the number of times the K-Tokeniser selects either the BERT Tokeniser or the Drug Tokeniser out of the three tokenisers. We find that the accuracy of the K-Tokeniser exceeds 99\% in recognising Drug concepts. However, it tends to make occasional mistakes by selecting the Disease Tokeniser even when a concept is labelled as either {\em B-Drug} or {\em I-Drug} in the dataset, as depicted in Table ~\ref{tab:06}. We observe that the K-Tokeniser consistently prefers a tokeniser that generates a lesser number of subwords among the three available options in the majority of cases. For instance, consider cases like {\em meropenem} and {\em donazol}, where the K-Tokeniser produces more meaningful subwords compared to the other two tokenisers. Despite this, such occurrences are deemed legitimate errors by the tokeniser since these subwords do not originate from either of the other two tokenisers.
\begin{table}[!ht]
\centering
\caption{Selection errors made by the K-Tokeniser} \label{tab:06}
\begin{tabular}{|l|l|l|l|}
\hline 
word & K-Tokeniser & BERT Tokeniser & Drug-Tokeniser \\
\hline 
chemo & [``ch", ``\#\#emo"] &[``ch", ``\#\#em", ``\#\#o"]&[``ch", ``\#\#em", ``\#\#o"] \\ 
meropenem& [``me", ``\#\#rop", ``\#\#enem"] & [``me", ``\#\#rop", ``\#\#ene", ``\#\#m"]&[``me", ``\#\#rope", ``\#\#ne", ``\#\#m"]\\
opioid &[``op", ``\#\#ioid"]& [``op", ``\#\#io", ``\#\#id"]&[``op", ``\#\#ioi", ``\#\#d"] \\ 
coags& [``co", ``\#\#ags"] &[``co", ``\#\#ag", ``\#\#s"]&[``co", ``\#\#ag", ``\#\#s"]\\
donazol& [``don", ``\#\#azol"] &[``don", ``\#\#az", ``\#\#ol"]&[``don", ``\#\#azo", ``\#\#l"]\\
\hline 
\end{tabular}
\vspace{0.2cm}
\caption*{The experiment is conducted using the n2c2 dataset with the K-Tokeniser based on the UMLS ontology. The first column lists the words, while the subsequent columns display the outputs from the K-Tokeniser, BERT Tokeniser, and Drug Tokeniser, respectively. It is important to note that the K-Tokeniser output differs from the other two tokenisers. Specifically, when the K-Tokeniser's output diverges from the other tokenisers in recognizing drug concepts, it is considered incorrect.
}
\end{table}
\par Finally, we analyse the performance at individual class levels on the ICD-9 automated coding tasks; see Figure ~\ref{fig:05}. We can observe that the overall trend of improvements of K-ClinicalBERT$_{UMLS}$ model over the baseline ClincalBERT is generally consistent and is not solely dependent on the label frequency in the training data. For the top 10 codes, we find that the K-ClinicalBERT$_{UMLS}$ model surpasses the baseline model in all cases; see Figure ~\ref{fig:05}(a). Notably, among the 10 most frequent codes in the dataset, the model achieves $F_1$ scores of 80\% for several codes, including 401.9 (Hypertension, NOS), 427.31 (Atrial fibrillation), and 414.01 (Coronary atherosclerosis of native coronary artery). In cases of the least frequent 10 codes, the ClinicalBERT$_{UMLS}$ always improves over the baseline model; see Figure ~\ref{fig:05}(b). Noteworthy mentions are 99.15 (Parenteral infusion of concentrated nutritional substances), 410.71 (Subendocardial infarction), 424.0 (Mitral valve disorders), 276.1 (Hyposmolality), and 995.92 (Severe Sepsis); see Figure ~\ref{fig:05}(b). We extend our analysis to find the performance improvements for disease related codes in order to evaluate the efficacy of semantic based tokenisation approach; see Figure ~\ref{fig:05}(c). Among the 10 categories of codes within the Disease chapter, the ClinicalBERT$_{UMLS}$ model achieves $F_1$ scores ranging from 60\% to 78\% for 6 of these categories. We further provide label-wise results achieved by ClinicalBERT and K-ClinicalBERT$_{UMLS}$ models for the top 50 labels in Supplementary Table 4 of \href{./supplementary_files.pdf}{Supplementary Files}. 
\par One of the limitations of our study is that we have not further pre-trained the models, as demonstrated in a recently published multilingual study \cite{dobler2023focus}. While their methodology initialises the embedding layer of a pre-trained model similar to ours, it does not incorporate the ontology-based tokenisation approach utilised in our research. We emphasise that in clinical settings, resource constraints may necessitate the use of smaller models. In such scenarios, our study demonstrates the effectiveness of increasing the model's capacity through augmentation of the tokeniser and the embedding layer without additional pre-training. In the future, we plan to expand our approach by integrating the knowledge graph approach into the embedding initialisation and further pre-training steps.
 \afterpage{\clearpage \begin{figure}[!h]
 \caption{\textbf{Comparison of classification results for individual labels (from top 50 ICD-9 codes) and disease chapters}}\label{fig:05}
 \begin{adjustbox}{left}
 \includegraphics[scale=0.35]{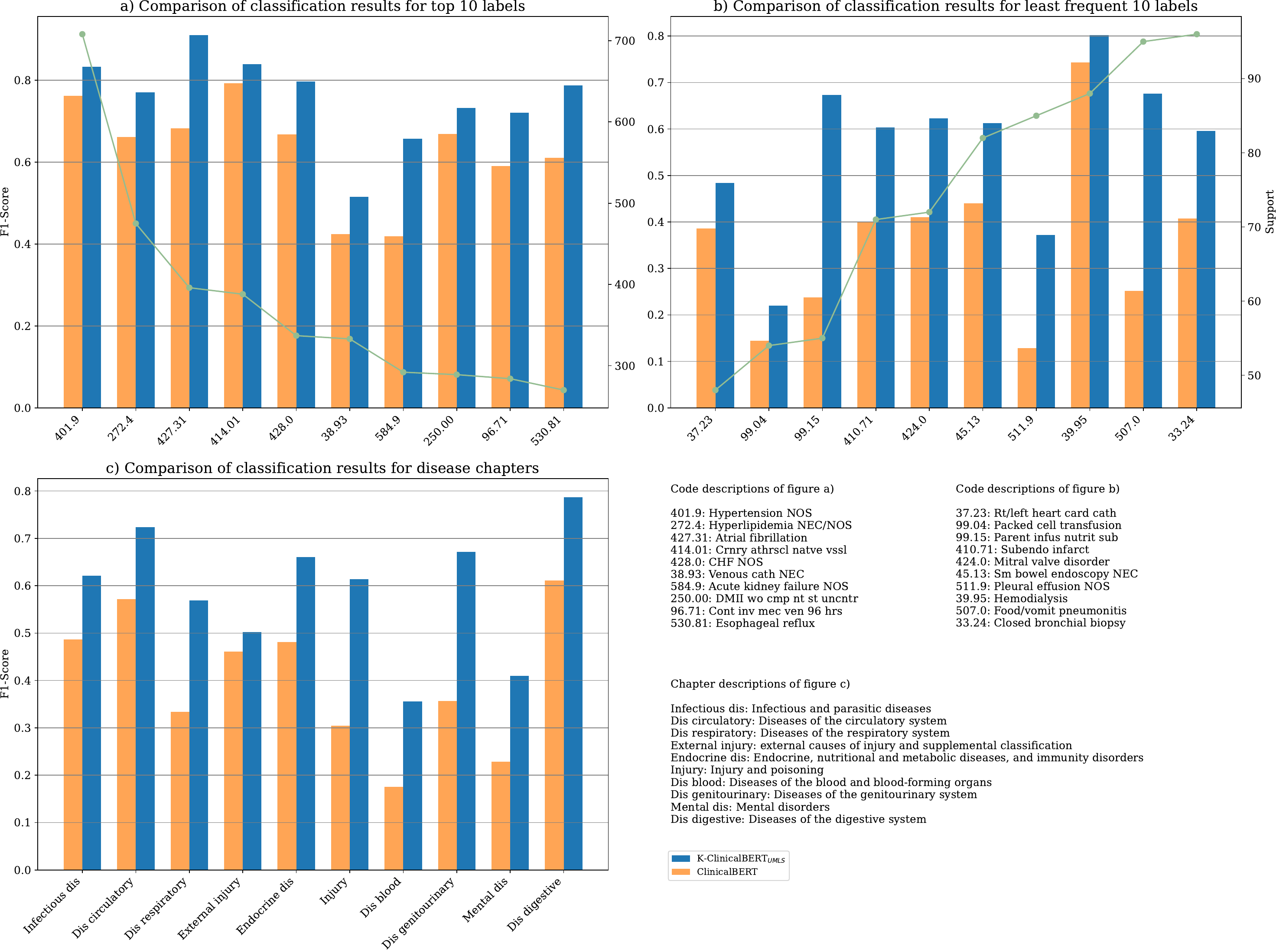}
 \end{adjustbox}
 \caption*{The results displayed include: a) the top 10 labels, b) the 10 least frequent labels, and c) labels associated with disease chapters. In the top row figures, the green line represents the support numbers for each label. The K-ClinicalBERT$_{UMLS}$ model is constructed using the K-Tokeniser from the UMLS corpus.} 
    
\end{figure}
\clearpage}
\section{Methods}
\label{sec:05}
\subsection*{Data Sources}
The primary data sources for this study are UMLS ontology, MIMIC-III, and PubMed corpora to generate sub-corpora for learning Drug and Disease vocabulary. In total, we collect 195519, 8555, and 58750 Drug related concepts from UMLS, MIMIC-III, and PubMed corpora, respectively. Similarly,  65511, 8859, and 24171 Disease related concepts are collected from three data sources, respectively. To generate the sub-corpora from MIMIC and PubMed corpora We perform minimal pre-processing including (i) removing illegal characters, and (ii) detecting word and sentence boundaries. For model development we load pre-trained weights of ClinicalBERT, PubMedBERT, and Bioformer models from HunggingFace \cite{wolf2019huggingface}. In the following subsections, we describe our K-Tokenisation methodology following the steps in Figure ~\ref{fig:01}.
\subsection*{K-Tokenisation}
\subsubsection*{Generating Sub-corpus by Semantic Type}
In the first step, medical concepts are extracted from the three data sources described above. Within the UMLS, these medical concepts are structured around specific semantic types. Our concept extraction methodologies from the UMLS and either corpus are facilitated through the utilisation of the UMLS MetamorphoSys \cite{bodenreider2004unified} and SemEHR \cite{wu2018semehr} tools, respectively. Specifically, we generate four types of semantic categories: (i) Pharmacologic Substance, (ii) Clinical Drug, (iii) Sign or Symptom, and (iv) Disease or Syndrome. The initial two semantic types, coalesce under the unified label of ``Drug'', while the subsequent pair are collectively labelled ``Disease''. The result of this process provides us with the concept frequencies, which are stored for further processing utilising tokenisation algorithms described next. 
\subsubsection*{Tokenisation Algorithms}
In the next step, we implement a knowledge-infused Byte Pair Encoding (BPE) algorithm (see Supplementary Algorithm 1 in \href{./supplementary_files.pdf}{Supplementary Files}) to expand the vocabulary of the BERT tokeniser. 
Our Supplementary Algorithm 2 in \href{./supplementary_files.pdf}{Supplementary Files} initially identifies all substrings within a semantic type Sub-corpus and calculates their frequencies (Line 5 in Supplementary Algorithm 2 of \href{./supplementary_files.pdf}{Supplementary Files}). For instance, in creating subwords related to Drugs, each concept is processed to gather all possible sub-strings or character n-grams and accumulate their frequencies from the Sub-corpus, 
which results in a corpus $C_d$. Subsequently, we process each concept in $C_d$ with a budget of $\alpha$ for the number of subwords to be produced by the BPE algorithm (i.e. Supplementary Algorithm 1 of \href{./supplementary_files.pdf}{Supplementary Files}); we set $\alpha$ to 20 for all the cases. This step generates a set of subwords for each concept that are candidates for merge operations. Similar to the decoding process in BERT tokenisers, concepts are not decoded based on the original order of merges but via a greedy largest subsequence left-to-right inference (See Line 9 in Supplementary Algorithm 2 of \href{./supplementary_files.pdf}{Supplementary Files}). However it sometimes produces lengthy prefixes and suffixes, which are further segmented into syllables (we apply syllable rules to the prefixes of these subwords if they exceed a length of 4). The rest of the subwords are discarded while keeping those found by the decoding process. Thus, a collection of subwords is gathered which represents the global character representation of Drug semantic type. We process Disease concepts in a similar way to construct a corpus, $C_s$, and relevant subwords. Finally, we collate all the subword frequencies into a single corpus $C$.  
\subsubsection*{K-Tokeniser Pool}
\begin{figure}[!h]
\centering
\caption{\textbf{The optimisation steps performed by the K-Tokeniser at both the word and sentence levels}} \label{fig:06}
   \includegraphics[width=.90\textwidth]{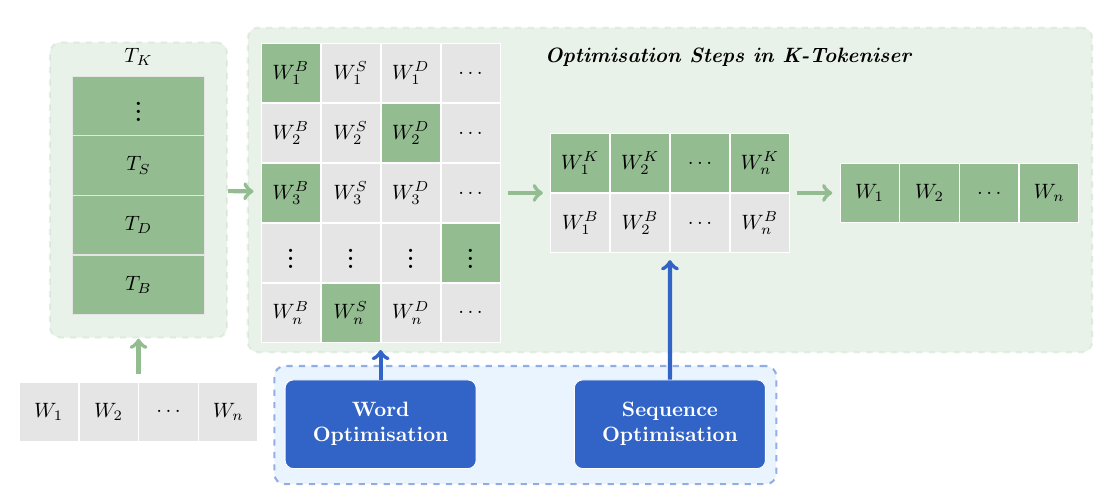}
 \caption*{A sentence composed of words $W_1, W_2, \ldots, W_n$ is input to the K-Tokeniser, $T_K$ . Various subword sequences generated by the $T_K$ tokeniser are organized in a matrix, with columns representing subwords from $T_B$, $T_S$, and $T_D$. Here, $T_B$, $T_S$, and $T_D$ denote BERT Tokeniser, Drug Tokeniser, and Disease Tokeniser, respectively. Dots in the tokeniser pool and matrix represent that this method can be extended to any number of semantic types. The word optimisation step selects the most suitable column for each word. Next, the sentence optimisation step compares the fertility between the default tokeniser(i.e. BERT Tokeniser in this case) and the K-Tokeniser.}
\end{figure}

After generating subwords for different semantic types, we augment them with the BERT Tokeniser. Specifically, we construct three distinct tokenisers: (i) BERT Tokeniser, denoted as $\mathrm{T}_{B}$; 
(ii) $\mathrm{T}_{B}$ augmented with Drug related subwords, denoted as $\mathrm{T}_{D}$; (iii) $\mathrm{T}_{B}$ augmented with Disease related subwords, denoted as $\mathrm{T}_{S}$. Therefore, our final tokeniser is a pool of tokenisers, which we call {\em K-Tokeniser} and denote it as $\mathrm{T}_{K}$; see Figure ~\ref{fig:06}. The K-Tokeniser can formally be defined as $\mathrm{T}_{K} = \mathrm{T}_{B} \cup \mathrm{T}_{D} \cup \mathrm{T}_{S}$; where $\mathrm{T}_{B}$ is also termed as {\em default tokeniser}. This construction leads to multiple types of segmentations for a single word. For instance, a word could be segmented differently by each $\mathrm{T}_{B}$, $\mathrm{T}_{D}$, and $\mathrm{T}_{S}$, resulting in three separate subword sequences. To address this, we have devised global and local optimisation steps, described later in our methodology.  
\subsubsection*{Embedding Initialiser}
The new subwords introduced in the K-Tokeniser typically lack representation in the pre-trained Transformer model. A common approach involves initialising the embedding layer of the pre-trained model randomly and subsequently fine-tuning it for the downstream task. However, this strategy might lead to a performance drop when the added vocabulary size is extensive and contains few examples in the downstream dataset. To mitigate this challenge, we create a back-off dictionary that maps K-Tokeniser subwords, i.e. subwords in $\mathrm{T}_{K}$  to the default tokeniser, $\mathrm{T}_{B}$. The mapping process involves executing the $\mathrm{T}_{B}$ on the newly introduced subword vocabulary of the $\mathrm{T}_{K}$ to determine their corresponding subwords. Let $t_i^k$ denotes a subword in $\mathrm{T}_{K}$, and $t_j^b$ denotes a subword in $\mathrm{T}_{B}$. We tokenise $t_i^k$ using $\mathrm{T}_{B}$, generating subwords $t_1^b, t_2^b, \ldots, t_l^b$,  where $l$ is the number of subwords. We initialise the embedding layer of a pre-trained BERT model by the averaging embeddings of $t_1^b, t_2^b, \ldots,t_l^b$. This procedure ensures the infusion of knowledge into the embedding layer through the K-Tokeniser, enhancing its adaptability for downstream tasks. 
\subsection*{Optimisation Steps in K-Tokeniser}
In the following subsections, we describe word (global) and sentence-level (local) optimisation procedures executed by the K-Tokeniser in scenarios involving three tokenisers. A schematic of the optimisation steps is outlined in Figure ~\ref{fig:06}.
\subsubsection*{Word Optimisation}
In the K-Tokeniser, each word is tokenised using $\mathrm{T}_{B}$, $\mathrm{T}_{D}$, and $\mathrm{T}_{S}$ independently, resulting in three different types of subwords. Formally, for a word $W_i$ from a sequence of words $W_1, W_2, W_3, \ldots, W_n$ of length $n$, the $\mathrm{T}_{K}$ tokeniser produces subwords $W_i^B$, $W_i^D$, and $W_i^S$ (see the matrix in Figure~\ref{fig:06}). These subwords may have different lengths. To select appropriate semantic based representation for the word $W_i$, we employ an entropy minimisation procedure. It selects the character representation that best matches the minimum entropy computed from the corpus $C$ originated after the execution of the Tokenisation Algorithms. Let the cumulative probabilities of subwords for $W_i^B$, $W_i^D$, and $W_i^S$ be denoted as $p_i^B$, $p_i^D$, and $p_i^S$, respectively, calculated from $C$. Then, the entropy $H_i^X$ for each type of subword is given by:
\begin{equation}
\centering
H_i^X = p_i^X \log p_i^X, \quad \text{where} \quad X \in {B, D, S}
\end{equation}
The objective of the K-tokeniser is to select subwords that minimise the entropy, $H_i^K$, i.e.,
\begin{equation}
\centering
H_i^K = \min\{H_i^B, H_i^D, H_i^S\}
\end{equation}
Thus, this approach ensures the selection of subwords with the least entropy, optimising the global character representation based on different semantic types.
However, in cases where both $H_i^{D}$ and $H_i^{S}$ fail to meet the threshold criteria (set to the minimum frequency of 1000), the minimum is  $H_i^{B}$ which selects segmentation obtained using $\mathrm{T}_{B}$. The procedure is repeated for the whole sequence or sentence and at the end of this iteration we obtain the the sequence, $W_1^K, W_2^K, \ldots W_n^K$, generated by the K-Tokeniser; see Figure ~\ref{fig:06}. Subsequently, a sentence-level optimisation is conducted to determine the appropriateness of the chosen segmentation. 
\subsubsection*{Sequence Optimisation}
The semantic based word-level optimisation step described above typically produces shorter sequences that are not seen by a model during its pre-training stage. To counterbalance this effect, we need a metric that preserves the tokeniser's knowledge while enhancing the model's performance. With this objective, we quantify the number of subwords produced by the K-Tokeniser, $\mathrm{T}_{K}$, and compare it to the default tokeniser, $\mathrm{T}_{B}$. Formally, if $f_i^{B}$ denotes the fertility (number of subwords produced by the tokeniser for a word $i$) of the default tokeniser, then its sentence fertility, $F_B$, can be written as:
\begin{equation}
    F^{B} = \sum_{i=1}^{n} f_i^{B}
\end{equation}
where $n$ is the length of the sentence/sequence. Similarly, the sentence fertility of the K-Tokeniser is:
\begin{equation}
    F^{K} = \sum_{i=1}^{n} f_i^{K}
\end{equation}
We calculate the fertility threshold, $F_{th}$, using the following formula:
\begin{equation}
    F_{th} = \frac{F^{B} - F^{K}}{F^{B}}
\end{equation}
The $F_{th}$ ratio provides a metric to adjust at the sentence level along with some predefined threshold values (i.e. 0, 1, 0.035, and 0.065 as discussed in the Discussion section). When the $F_{th}$ ratio surpasses the specified threshold, the K-Tokeniser disregards its outputs and reverts to generating outputs from the  default tekeniser (i.e. BERT Tokeniser in this case). This mechanism governs the K-Tokeniser's behaviour in a local context, ensuring it switches to the original tokeniser when it generates fewer subwords than required.


\subsection*{Evaluation of K-Tokeniser}
In our evaluation, we assess BERT models using Micro $F_1$ scores across all tasks. Additionally, we provide overall precision and recall scores; see Supplementary Table 1, 2, and 3 in \href{./supplementary_files.pdf}{Supplementary Files}. For the automated clinical coding task, we report both Macro and Micro precision and recall scores. 
\subsubsection*{Fine tune ClinicalBERT for clinical concepts and phenotype identification}
 We approach the clinical concepts and phenotype identification tasks as sequence labelling tasks. The datasets for these tasks follow the ``BIO'' labelling schema, where ``B-'' and ``I-'' prefixes denote words at the beginning and inside of a concept, and ``O'' stands for words located outside of any concepts of interest. Each word in the dataset is annotated with one of the BIO labels. The input to the ClinicalBERT model consists of batches of sentences and their corresponding word labels. A linear classifier layer is added to compute the probability of each word belonging to a category. Training involves fine-tuning Transformer layers to produce hidden representations for each word and then passing them to the linear classifier layer to compute the cross-entropy loss. We utilise the Hugging Face Transformers library \cite{wolf2019huggingface} for fine-tuning the ClinicalBERT model.
\subsubsection*{Fine tune ClinicalBERT for clinical relation extraction}
We approach clinical relation extraction as a classification task. Following previous studies \cite{yang2020identifying, yang2022large}, each relation is identified with two sets of entity markers where [S1], [E1] denote the first sentence and first entity, and [S2] and [E2] denote the second sentence and second entity. The two sentences where there is a relation are placed one after another and use a special token ([SEP]) to identify their boundary. The sentences are processed by concatenating the representations of the model's special [CLS] token and all four entity markers \cite{yang2020identifying}. A classification layer is added for classification and the cross-entropy loss is used to fine-tune ClinicalBERT models.
\subsubsection*{Fine tune ClinicalBERT for automated clinical coding}
We adopt the state-of-the-art PLM-ICD architecture proposed by Huang et al. \cite{huang2022plm} for the task of automated clinical coding. In this architecture, documents are first divided into segments with a chunk size of 128. Each segment is then separately inputted into the ClinicalBERT layer to compute hidden representations. The [CLS] representations of each segment are concatenated to obtain hidden representations of all words in the document. Subsequently, attention scores are calculated for each label in the dataset for the document in question, which are then used to compute hidden representations of the document. Finally, a sigmoid layer is employed for document classification \cite{huang2022plm}.  
\subsubsection*{Fine tune PubMedBERT and Bioformer for clinical research article classification}
The task of clinical research article classification involves multi-class classification, where both the PubMedBERT and Bioformer models are enhanced with a linear classifier layer. If the document length exceeds the maximum limit, any remaining tokens are discarded. The input document is then fed into the PubMedBERT model to compute the hidden representation of the [CLS] token, which serves as the document representation. This representation is subsequently passed to the classifier layer to compute the cross-entropy loss during fine-tuning. A similar approach is followed for fine-tuning using the Bioformer model.
\subsection*{Training Details}
Our models are trained on the cross-entropy loss function with the AdamW optimizer, wherein the learning rate is set to $2 \times 10^{-5}$ and the weight decay to 0.01. Specifically, we fine-tune the models for the clinical relation extraction task for 3 epochs and automated clinical coding tasks for 20 epochs, while all other models tailored for distinct tasks are fine-tuned for 10 epochs. The batch sizes are uniformly set to 8 for all tasks, with the exception of the automated clinical coding task, where a batch size of 2 is employed. All experiments are conducted using a single NVIDIA RTX A5000 GPU equipped with 24GB of graphics memory.
\section*{Data availability}
To generate sub corpora for K-Tokeniser we utilise MIMIC-III, UMLS ontology, and PubMed abstracts. We pre-process MIMIC-III and PubMed corpora using the SemEHR tool. They can be downloaded from the following addresses:  
\begin{enumerate}
    \item Access to MIMIC-III can be requested at \url{https://physionet.org/content/mimiciii/}, which requires a signed safe usage agreement.
    \item Instructions for accessing the UMLS MetamorphoSys tool can be found at \url{https://www.nlm.nih.gov/research/umls/index.html}.
    \item PubMed data can be downloaded from \url{https://pubmed.ncbi.nlm.nih.gov/download/}.
    \item The SemEHR tool can be downloaded from \url{https://github.com/CogStack/CogStack-SemEHR}.
\end{enumerate}

\section*{Code availability}
\begin{enumerate}
    \item The code for automated coding classification using the PLM-ICD architecture can be found at: \url {https://github.com/MiuLab/PLM-ICD}. Our code integrating K-Tokeniser is in: 
    \item The code for clinical relation extraction used in a previous study \cite{yang2022large} can be found in; \url{https://github.com/uf-hobi-informatics-lab/ClinicalTransformerRelationExtraction}. Our code integrating K-tokeniser is in: \url{}
    \item All our codes will be published in: \url{https://github.com/knowlab/KTokenization} 
    
\end{enumerate}
 \section*{Acknowledgement}
This work was supported by UK’s Medical Research Council (MR/S004149/1, MR/X030075/1); National Institute for Health Research (NIHR202639); British Council (UCL-NMU-SEU International Collaboration On Artificial Intelligence In Medicine: Tackling Challenges Of Low Generalisability And Health Inequality) and (Facilitating Better Urology Care With Effective And Fair Use Of Artificial Intelligence - A Partnership Between UCL And Shanghai Jiao Tong University School Of Medicine); HW’s role in this research was partially funded by the Legal \& General Group (research grant to establish the independent Advanced Care Research Centre at University of Edinburgh). The funders had no role in conduct of the study, interpretation, or the decision to submit for publication. The views expressed are those of the authors and not necessarily those of Legal \& General.
\section*{Contributions}
All authors contributed to study design. AH and HW were involved in K-Tokeniser algorithm design and development. AH coded the K-Tokeniser, JW and QN helped in running downstream tasks. AH drafted manuscript, HW contributed significantly. All authors provided revisions, and final approval.
\section*{Competing interests}
The authors declare no competing interests.
\bibliography{library}

\begin{thebibliography}{10}
\urlstyle{rm}
\expandafter\ifx\csname url\endcsname\relax
  \def\url#1{\texttt{#1}}\fi
\expandafter\ifx\csname urlprefix\endcsname\relax\def\urlprefix{URL }\fi
\expandafter\ifx\csname doiprefix\endcsname\relax\def\doiprefix{DOI: }\fi
\providecommand{\bibinfo}[2]{#2}
\providecommand{\eprint}[2][]{\url{#2}}

\bibitem{devlin2019bert}
\bibinfo{author}{Devlin, J.}, \bibinfo{author}{Chang, M.-W.},
  \bibinfo{author}{Lee, K.} \& \bibinfo{author}{Toutanova, K.}
\newblock \bibinfo{title}{{BERT: Pre-training of Deep Bidirectional
  Transformers for Language Understanding}}.
\newblock In \emph{\bibinfo{booktitle}{{Proceedings of the 2019 Conference of
  the North American Chapter of the Association for Computational Linguistics:
  Human Language Technologies, Volume 1 (Long and Short Papers)}}},
  \bibinfo{pages}{4171--4186} (\bibinfo{year}{2019}).

\bibitem{brown2020language}
\bibinfo{author}{Brown, T.} \emph{et~al.}
\newblock \bibinfo{journal}{\bibinfo{title}{{Language Models are Few-Shot
  Learners}}}.
\newblock {\emph{\JournalTitle{{Advances in Neural Information Processing
  Systems}}}} \textbf{\bibinfo{volume}{33}}, \bibinfo{pages}{1877--1901}
  (\bibinfo{year}{2020}).

\bibitem{achiam2023gpt}
\bibinfo{author}{Achiam, J.} \emph{et~al.}
\newblock \bibinfo{journal}{\bibinfo{title}{{GPT-4 Technical Report}}}.
\newblock {\emph{\JournalTitle{{arXiv preprint arXiv:2303.08774}}}}
  (\bibinfo{year}{2023}).

\bibitem{vaswani2017attention}
\bibinfo{author}{Vaswani, A.} \emph{et~al.}
\newblock \bibinfo{journal}{\bibinfo{title}{{Attention Is All You Need}}}.
\newblock {\emph{\JournalTitle{{Advances in Neural Information Processing
  Systems}}}} \textbf{\bibinfo{volume}{30}} (\bibinfo{year}{2017}).

\bibitem{yehezkel2023incorporating}
\bibinfo{author}{Yehezkel, S.} \& \bibinfo{author}{Pinter, Y.}
\newblock \bibinfo{title}{{Incorporating Context into Subword Vocabularies}}.
\newblock In \emph{\bibinfo{booktitle}{Proceedings of the 17th Conference of
  the European Chapter of the Association for Computational Linguistics}},
  \bibinfo{pages}{623--635} (\bibinfo{year}{2023}).

\bibitem{alsentzer2019publicly}
\bibinfo{author}{Alsentzer, E.} \emph{et~al.}
\newblock \bibinfo{journal}{\bibinfo{title}{{Publicly Available Clinical {BERT}
  Embeddings}}}.
\newblock {\emph{\JournalTitle{{NAACL HLT 2019}}}} \bibinfo{pages}{72}
  (\bibinfo{year}{2019}).

\bibitem{gu2021domain}
\bibinfo{author}{Gu, Y.} \emph{et~al.}
\newblock \bibinfo{journal}{\bibinfo{title}{{Domain-Specific Language Model
  Pretraining for Biomedical Natural Language Processing}}}.
\newblock {\emph{\JournalTitle{{ACM Transactions on Computing for Healthcare
  (HEALTH)}}}} \textbf{\bibinfo{volume}{3}}, \bibinfo{pages}{1--23}
  (\bibinfo{year}{2021}).

\bibitem{yang2022large}
\bibinfo{author}{Yang, X.} \emph{et~al.}
\newblock \bibinfo{journal}{\bibinfo{title}{A large language model for
  electronic health records}}.
\newblock {\emph{\JournalTitle{npj digital medicine}}}
  \textbf{\bibinfo{volume}{5}}, \bibinfo{pages}{194} (\bibinfo{year}{2022}).

\bibitem{lee2020biobert}
\bibinfo{author}{Lee, J.} \emph{et~al.}
\newblock \bibinfo{journal}{\bibinfo{title}{{BioBERT: a pre-trained biomedical
  language representation model for biomedical text mining}}}.
\newblock {\emph{\JournalTitle{Bioinformatics}}} \textbf{\bibinfo{volume}{36}},
  \bibinfo{pages}{1234--1240} (\bibinfo{year}{2020}).

\bibitem{tai2020exbert}
\bibinfo{author}{Tai, W.}, \bibinfo{author}{Kung, H.}, \bibinfo{author}{Dong,
  X.~L.}, \bibinfo{author}{Comiter, M.} \& \bibinfo{author}{Kuo, C.-F.}
\newblock \bibinfo{title}{{exBERT: Extending pre-trained models with
  domain-specific vocabulary under constrained training resources}}.
\newblock In \emph{\bibinfo{booktitle}{Findings of the Association for
  Computational Linguistics: EMNLP 2020}}, \bibinfo{pages}{1433--1439}
  (\bibinfo{year}{2020}).

\bibitem{rust2021good}
\bibinfo{author}{Rust, P.}, \bibinfo{author}{Pfeiffer, J.},
  \bibinfo{author}{Vuli{\'c}, I.}, \bibinfo{author}{Ruder, S.} \&
  \bibinfo{author}{Gurevych, I.}
\newblock \bibinfo{title}{{How Good is Your Tokenizer? On the Monolingual
  Performance of Multilingual Language Models}}.
\newblock In \emph{\bibinfo{booktitle}{{Proceedings of the 59th Annual Meeting
  of the Association for Computational Linguistics and the 11th International
  Joint Conference on Natural Language Processing (Volume 1: Long Papers)}}},
  \bibinfo{pages}{3118--3135} (\bibinfo{year}{2021}).

\bibitem{bodenreider2004unified}
\bibinfo{author}{Bodenreider, O.}
\newblock \bibinfo{journal}{\bibinfo{title}{{The unified medical language
  system (UMLS): integrating biomedical terminology}}}.
\newblock {\emph{\JournalTitle{{Nucleic acids research}}}}
  \textbf{\bibinfo{volume}{32}}, \bibinfo{pages}{D267--D270}
  (\bibinfo{year}{2004}).

\bibitem{johnson2016mimic}
\bibinfo{author}{Johnson, A.~E.} \emph{et~al.}
\newblock \bibinfo{journal}{\bibinfo{title}{{MIMIC-III, a freely accessible
  critical care database}}}.
\newblock {\emph{\JournalTitle{scientific data}}} \textbf{\bibinfo{volume}{3}},
  \bibinfo{pages}{1--9} (\bibinfo{year}{2016}).

\bibitem{sennrich2016neural}
\bibinfo{author}{Sennrich, R.}, \bibinfo{author}{Haddow, B.} \&
  \bibinfo{author}{Birch, A.}
\newblock \bibinfo{title}{{Neural Machine Translation of Rare Words with
  Subword Units}}.
\newblock In \emph{\bibinfo{booktitle}{Proceedings of the 54th Annual Meeting
  of the Association for Computational Linguistics (Volume 1: Long Papers)}},
  \bibinfo{pages}{1715--1725} (\bibinfo{year}{2016}).

\bibitem{schuster2012japanese}
\bibinfo{author}{Schuster, M.} \& \bibinfo{author}{Nakajima, K.}
\newblock \bibinfo{title}{Japanese and korean voice search}.
\newblock In \emph{\bibinfo{booktitle}{2012 IEEE international conference on
  acoustics, speech and signal processing (ICASSP)}},
  \bibinfo{pages}{5149--5152} (\bibinfo{organization}{IEEE},
  \bibinfo{year}{2012}).

\bibitem{kudo2018sentencepiece}
\bibinfo{author}{Kudo, T.} \& \bibinfo{author}{Richardson, J.}
\newblock \bibinfo{title}{{SentencePiece: A simple and language independent
  subword tokenizer and detokenizer for Neural Text Processing}}.
\newblock In \emph{\bibinfo{booktitle}{Proceedings of the 2018 Conference on
  Empirical Methods in Natural Language Processing: System Demonstrations}},
  \bibinfo{pages}{66--71} (\bibinfo{year}{2018}).

\bibitem{mielke2021between}
\bibinfo{author}{Mielke, S.~J.} \emph{et~al.}
\newblock \bibinfo{journal}{\bibinfo{title}{Between words and characters: a
  brief history of open-vocabulary modeling and tokenization in nlp}}.
\newblock {\emph{\JournalTitle{arXiv preprint arXiv:2112.10508}}}
  (\bibinfo{year}{2021}).

\bibitem{xu2021vocabulary}
\bibinfo{author}{Xu, J.}, \bibinfo{author}{Zhou, H.}, \bibinfo{author}{Gan,
  C.}, \bibinfo{author}{Zheng, Z.} \& \bibinfo{author}{Li, L.}
\newblock \bibinfo{title}{{Vocabulary Learning via Optimal Transport for Neural
  Machine Translation}}.
\newblock In \emph{\bibinfo{booktitle}{Proceedings of the 59th Annual Meeting
  of the Association for Computational Linguistics and the 11th International
  Joint Conference on Natural Language Processing (Volume 1: Long Papers)}},
  \bibinfo{pages}{7361--7373} (\bibinfo{year}{2021}).

\bibitem{kudo2018subword}
\bibinfo{author}{Kudo, T.}
\newblock \bibinfo{title}{{Subword Regularization: Improving Neural Network
  Translation Models with Multiple Subword Candidates}}.
\newblock In \emph{\bibinfo{booktitle}{Proceedings of the 56th Annual Meeting
  of the Association for Computational Linguistics (Volume 1: Long Papers)}},
  \bibinfo{pages}{66--75} (\bibinfo{year}{2018}).

\bibitem{hofmann2022embarrassingly}
\bibinfo{author}{Hofmann, V.}, \bibinfo{author}{Sch{\"u}tze, H.} \&
  \bibinfo{author}{Pierrehumbert, J.}
\newblock \bibinfo{title}{{An Embarrassingly Simple Method to Mitigate
  Undesirable Properties of Pretrained Language Model Tokenizers}}.
\newblock In \emph{\bibinfo{booktitle}{Proceedings of the 60th Annual Meeting
  of the Association for Computational Linguistics (Volume 2: Short Papers)}},
  \bibinfo{pages}{385--393} (\bibinfo{year}{2022}).

\bibitem{hong2021avocado}
\bibinfo{author}{Hong, J.}, \bibinfo{author}{Kim, T.}, \bibinfo{author}{Lim,
  H.} \& \bibinfo{author}{Choo, J.}
\newblock \bibinfo{title}{{AVocaDo: Strategy for Adapting Vocabulary to
  Downstream Domain}}.
\newblock In \emph{\bibinfo{booktitle}{{Proceedings of the 2021 Conference on
  Empirical Methods in Natural Language Processing}}},
  \bibinfo{pages}{4692--4700} (\bibinfo{year}{2021}).

\bibitem{mosin2023fine}
\bibinfo{author}{Mosin, V.}, \bibinfo{author}{Samenko, I.},
  \bibinfo{author}{Kozlovskii, B.}, \bibinfo{author}{Tikhonov, A.} \&
  \bibinfo{author}{Yamshchikov, I.~P.}
\newblock \bibinfo{journal}{\bibinfo{title}{{Fine-tuning transformers:
  Vocabulary transfer}}}.
\newblock {\emph{\JournalTitle{Artificial Intelligence}}}
  \textbf{\bibinfo{volume}{317}}, \bibinfo{pages}{103860}
  (\bibinfo{year}{2023}).

\bibitem{bostrom2020byte}
\bibinfo{author}{Bostrom, K.} \& \bibinfo{author}{Durrett, G.}
\newblock \bibinfo{title}{{Byte Pair Encoding is Suboptimal for Language Model
  Pretraining}}.
\newblock In \emph{\bibinfo{booktitle}{{Findings of the Association for
  Computational Linguistics: EMNLP 2020}}}, \bibinfo{pages}{4617--4624}
  (\bibinfo{year}{2020}).

\bibitem{xue2022byt5}
\bibinfo{author}{Xue, L.} \emph{et~al.}
\newblock \bibinfo{journal}{\bibinfo{title}{{ByT5: Towards a Token-Free Future
  with Pre-trained Byte-to-Byte Models}}}.
\newblock {\emph{\JournalTitle{Transactions of the Association for
  Computational Linguistics}}} \textbf{\bibinfo{volume}{10}},
  \bibinfo{pages}{291--306} (\bibinfo{year}{2022}).

\bibitem{henry20202018}
\bibinfo{author}{Henry, S.}, \bibinfo{author}{Buchan, K.},
  \bibinfo{author}{Filannino, M.}, \bibinfo{author}{Stubbs, A.} \&
  \bibinfo{author}{Uzuner, O.}
\newblock \bibinfo{journal}{\bibinfo{title}{{2018 n2c2 shared task on adverse
  drug events and medication extraction in electronic health records}}}.
\newblock {\emph{\JournalTitle{{Journal of the American Medical Informatics
  Association}}}} \textbf{\bibinfo{volume}{27}}, \bibinfo{pages}{3--12}
  (\bibinfo{year}{2020}).

\bibitem{yang2020identifying}
\bibinfo{author}{Yang, X.} \emph{et~al.}
\newblock \bibinfo{journal}{\bibinfo{title}{Identifying relations of
  medications with adverse drug events using recurrent convolutional neural
  networks and gradient boosting}}.
\newblock {\emph{\JournalTitle{Journal of the American Medical Informatics
  Association}}} \textbf{\bibinfo{volume}{27}}, \bibinfo{pages}{65--72}
  (\bibinfo{year}{2020}).

\bibitem{huang2022plm}
\bibinfo{author}{Huang, C.-W.}, \bibinfo{author}{Tsai, S.-C.} \&
  \bibinfo{author}{Chen, Y.-N.}
\newblock \bibinfo{journal}{\bibinfo{title}{{PLM-ICD: Automatic ICD Coding with
  Pretrained Language Models}}}.
\newblock {\emph{\JournalTitle{ClinicalNLP 2022}}} \bibinfo{pages}{10}
  (\bibinfo{year}{2022}).

\bibitem{mullenbach2018explainable}
\bibinfo{author}{Mullenbach, J.}, \bibinfo{author}{Wiegreffe, S.},
  \bibinfo{author}{Duke, J.}, \bibinfo{author}{Sun, J.} \&
  \bibinfo{author}{Eisenstein, J.}
\newblock \bibinfo{title}{{Explainable Prediction of Medical Codes from
  Clinical Text}}.
\newblock In \emph{\bibinfo{booktitle}{Proceedings of the 2018 Conference of
  the North American Chapter of the Association for Computational Linguistics:
  Human Language Technologies, Volume 1 (Long Papers)}},
  \bibinfo{pages}{1101--1111} (\bibinfo{year}{2018}).

\bibitem{dong2021explainable}
\bibinfo{author}{{Dong, Hang and Su{\'a}rez-Paniagua, V{\'\i}ctor and Whiteley,
  William and Wu, Honghan}}.
\newblock \bibinfo{journal}{\bibinfo{title}{{Explainable automated coding of
  clinical notes using hierarchical label-wise attention networks and label
  embedding initialisation}}}.
\newblock {\emph{\JournalTitle{{Journal of Biomedical Informatics}}}}
  \textbf{\bibinfo{volume}{116}}, \bibinfo{pages}{103728}
  (\bibinfo{year}{2021}).

\bibitem{jainradgraph}
\bibinfo{author}{{Jain, Saahil and Agrawal, Ashwin and Saporta, Adriel and
  Truong, Steven QH and Duong, Du Nguyen and Bui, Tan and Chambon, Pierre and
  Zhang, Yuhao and Lungren, Matthew P and Ng, Andrew Y and others}}.
\newblock \bibinfo{journal}{\bibinfo{title}{{RadGraph: Extracting Clinical
  Entities and Relations from Radiology Reports}}}.
\newblock {\emph{\JournalTitle{{arXiv preprint arXiv:2106.14463}}}}
  (\bibinfo{year}{2021}).

\bibitem{chen2022multi}
\bibinfo{author}{Chen, Q.} \emph{et~al.}
\newblock \bibinfo{journal}{\bibinfo{title}{{Multi-label classification for
  biomedical literature: an overview of the BioCreative VII LitCovid Track for
  COVID-19 literature topic annotations}}}.
\newblock {\emph{\JournalTitle{Database}}} \textbf{\bibinfo{volume}{2022}},
  \bibinfo{pages}{baac069} (\bibinfo{year}{2022}).

\bibitem{fang2023bioformer}
\bibinfo{author}{Fang, L.}, \bibinfo{author}{Chen, Q.}, \bibinfo{author}{Wei,
  C.-H.}, \bibinfo{author}{Lu, Z.} \& \bibinfo{author}{Wang, K.}
\newblock \bibinfo{journal}{\bibinfo{title}{{Bioformer: an efficient
  transformer language model for biomedical text mining}}}.
\newblock {\emph{\JournalTitle{arXiv preprint arXiv:2302.01588}}}
  (\bibinfo{year}{2023}).

\bibitem{de1998hierarchical}
\bibinfo{author}{De~Lima, L.~R.}, \bibinfo{author}{Laender, A.~H.} \&
  \bibinfo{author}{Ribeiro-Neto, B.~A.}
\newblock \bibinfo{title}{{A Hierarchical Approach to the Automatic
  Categorization of Medical Documents}}.
\newblock In \emph{\bibinfo{booktitle}{{Proceedings of the Seventh
  International Conference on Information and Knowledge Management}}},
  \bibinfo{pages}{132--139} (\bibinfo{year}{1998}).

\bibitem{dong2022automated}
\bibinfo{author}{Dong, H.} \emph{et~al.}
\newblock \bibinfo{journal}{\bibinfo{title}{Automated clinical coding: what,
  why, and where we are?}}
\newblock {\emph{\JournalTitle{NPJ digital medicine}}}
  \textbf{\bibinfo{volume}{5}}, \bibinfo{pages}{159} (\bibinfo{year}{2022}).

\bibitem{johnson2019mimic}
\bibinfo{author}{Johnson, A.~E.} \emph{et~al.}
\newblock \bibinfo{journal}{\bibinfo{title}{{MIMIC-CXR, a de-identified
  publicly available database of chest radiographs with free-text reports}}}.
\newblock {\emph{\JournalTitle{scientific data}}} \textbf{\bibinfo{volume}{6}},
  \bibinfo{pages}{317} (\bibinfo{year}{2019}).

\bibitem{irvin2019chexpert}
\bibinfo{author}{Irvin, J.} \emph{et~al.}
\newblock \bibinfo{title}{{CheXpert: A Large Chest Radiograph Dataset with
  Uncertainty Labels and Expert Comparison}}.
\newblock In \emph{\bibinfo{booktitle}{{Proceedings of the AAAI Conference on
  Artificial Intelligence}}}, vol.~\bibinfo{volume}{33},
  \bibinfo{pages}{590--597} (\bibinfo{year}{2019}).

\bibitem{chen2021litcovid}
\bibinfo{author}{{Chen, Qingyu and Allot, Alexis and Lu, Zhiyong}}.
\newblock \bibinfo{journal}{\bibinfo{title}{{LitCovid: an open database of
  COVID-19 literature}}}.
\newblock {\emph{\JournalTitle{{Nucleic Acids Research}}}}
  \textbf{\bibinfo{volume}{49}}, \bibinfo{pages}{D1534--D1540}
  (\bibinfo{year}{2021}).

\bibitem{searle2020experimentals}
\bibinfo{author}{Searle, T.}, \bibinfo{author}{Ibrahim, Z.} \&
  \bibinfo{author}{Dobson, R.}
\newblock \bibinfo{title}{{Experimental Evaluation and Development of a
  Silver-Standard for the MIMIC-III Clinical Coding Dataset}}.
\newblock In \emph{\bibinfo{booktitle}{{Proceedings of the 19th SIGBioMed
  Workshop on Biomedical Language Processing}}}, \bibinfo{pages}{76--85}
  (\bibinfo{year}{2020}).

\bibitem{dobler2023focus}
\bibinfo{author}{Dobler, K.} \& \bibinfo{author}{De~Melo, G.}
\newblock \bibinfo{title}{{FOCUS: Effective Embedding Initialization for
  Monolingual Specialization of Multilingual Models}}.
\newblock In \emph{\bibinfo{booktitle}{{Proceedings of the 2023 Conference on
  Empirical Methods in Natural Language Processing}}},
  \bibinfo{pages}{13440--13454} (\bibinfo{year}{2023}).

\bibitem{wolf2019huggingface}
\bibinfo{author}{Wolf, T.} \emph{et~al.}
\newblock \bibinfo{journal}{\bibinfo{title}{{HuggingFace's Transformers:
  State-of-the-art Natural Language Processing}}}.
\newblock {\emph{\JournalTitle{arXiv preprint arXiv:1910.03771}}}
  (\bibinfo{year}{2019}).

\bibitem{wu2018semehr}
\bibinfo{author}{Wu, H.} \emph{et~al.}
\newblock \bibinfo{journal}{\bibinfo{title}{{SemEHR: A general-purpose semantic
  search system to surface semantic data from clinical notes for tailored care,
  trial recruitment, and clinical research}}}.
\newblock {\emph{\JournalTitle{{Journal of the American Medical Informatics
  Association}}}} \textbf{\bibinfo{volume}{25}}, \bibinfo{pages}{530--537}
  (\bibinfo{year}{2018}).

\end{thebibliography}
\appendix
\label{apex01}
\section*{\href{./supplementary_files.pdf}
{Supplementary Files}}

\end{document}